\documentclass[format=acmsmall, review=false, screen=true]{acmart}

\makeatletter
\def\runningfoot{\def\@runningfoot{}}
\def\firstfoot{\def\@firstfoot{}}
\def\acmJournal{\def\@acmJournal{}}
\makeatother

\renewcommand\footnotetextcopyrightpermission[1]{} 
\usepackage{booktabs} 

\usepackage[ruled]{algorithm2e} 

\SetAlFnt{\small}
\SetAlCapFnt{\small}
\SetAlCapNameFnt{\small}
\SetAlCapHSkip{0pt}
\IncMargin{-\parindent}

\usepackage{url}
\usepackage{latexsym}
\usepackage{amsmath}
\usepackage{graphicx}
\usepackage[subfigure]{graphfig}
\usepackage{multirow}
\usepackage{footnote}
\def\runningfoot{\def\@runningfoot{}}
\def\firstfoot{\def\@firstfoot{}}
\begin{document}
\title[Familia: A Configurable Topic Modeling Framework for Industrial Text Engineering]{Familia: A Configurable Topic Modeling Framework for Industrial Text Engineering}
\titlenote{Some components of Familia have been open-sourced at Github (https://github.com/baidu/Familia).}
\titlenote{This is an extended and revised version of a preliminary paper that was presented in arXiv \cite{jiang2017familia}.}
\author{Di Jiang}\authornote{Both authors contributed equally to the paper}
\author{Yuanfeng Song}
\authornotemark[3]
\author{Rongzhong Lian}
\author{Siqi Bao}
\author{Jinhua Peng} 
\author{Huang He}
\author{Hua Wu}
\affiliation{
  \institution{Natural Language Processing Department, Baidu Inc.}
  \country{China}}
\email{ {jiangdi, songyuanfeng, lianrongzhong, baosiqi, pengjinhua, hehuang, wu_hua}@baidu.com }

\begin{abstract}
In the last decade, a variety of topic models have been proposed for text engineering. However, except Probabilistic Latent Semantic Analysis (PLSA) and Latent Dirichlet Allocation (LDA), most of existing topic models are seldom applied or considered in industrial scenarios. This phenomenon is caused by the fact that there are very few convenient tools to support these topic models so far. Intimidated by the demanding expertise and labor of designing and implementing parameter inference algorithms, software engineers are prone to simply resort to PLSA/LDA, without considering whether it is proper for their problem at hand or not.

In this paper, we propose a configurable topic modeling framework named Familia, in order to bridge the huge gap between academic research fruits and current industrial practice. Familia supports an important line of topic models that are widely applicable in text engineering scenarios. In order to relieve burdens of software engineers without knowledge of Bayesian networks, Familia is able to conduct automatic parameter inference for a variety of topic models. Simply through changing the data organization of Familia, software engineers are able to easily explore a broad spectrum of existing topic models or even design their own topic models, and find the one that best suits the problem at hand. With its superior extendability, Familia has a novel sampling mechanism that strikes balance between effectiveness and efficiency of parameter inference. Furthermore, Familia is essentially a big topic modeling framework that supports parallel parameter inference and distributed parameter storage. 
After being open-sourced at Github, Familia rapidly becomes the second most popular project in topic model area, and is widely used in both industrial community and academia.
The utilities and necessity of Familia are demonstrated in real-life industrial applications. Familia would significantly enlarge software engineers' arsenal of topic models and pave the way for utilizing highly customized topic models in real-life problems.
\end{abstract}



\maketitle

\renewcommand{\shortauthors}{D. Jiang et al.}

\section{Introduction}

Topic models have become one kind of important tools for text engineering. In the last decade, a wide spectrum of topic models has been proposed in academia and demonstrates promising performance. However, for industrial topic modeling, Probabilistic Latent Semantic Analysis (PLSA) \cite{hofmann1999probabilistic} and Latent Dirichlet Allocation (LDA) \cite{blei2003latent} are the working horses so far \cite{borisov2016using}\cite{si2010confucius}. With the richness of the other topic models, to name a few, TOT~\cite{wang2006topics}, Bilingual Topic Model~\cite{gao2011clickthrough}, Pair Model~\cite{jagarlamudi2013modeling}, GeoFolk~\cite{sizov2010geofolk}, LATM~\cite{wang2007mining} and Multifaceted Topic Model~\cite{vosecky2014integrating}, we rarely witness employment of them in industrial applications. For example, Bilingual Topic Model~\cite{gao2011clickthrough} is more suitable for modeling the query log data in search engine area, however, there are very few convenient tools to support it so far.

The huge gap between the abundance of topic models proposed in academia and their rare appearance in industry is mainly caused by the following reasons: 
\begin{enumerate}
\item Most of existing topic models do not have implementation for convenient usage. Implementing these topic models from scratch is both time-consuming and error-prone; 
\item Although many tasks cannot be suitably supported by existing topic models, designing a proper topic model and the corresponding parameter inference algorithms are daunting for engineers; 
\item Most advanced techniques for efficient parameter inference are exclusively designed for PLSA/LDA. Lacking highly efficient parameter inference algorithm impedes most topic models' applicability in industry.
\end{enumerate}

\noindent Due to the above reasons, engineers' topic modeling choice is usually limited to PLSA/LDA, which, however, may not fit well their task at hand. Such improper practice heavily undermines the effectiveness of topic models in real-life applications.

\begin{figure}[t!]
\begin{center}
\includegraphics[width=1.0\textwidth]{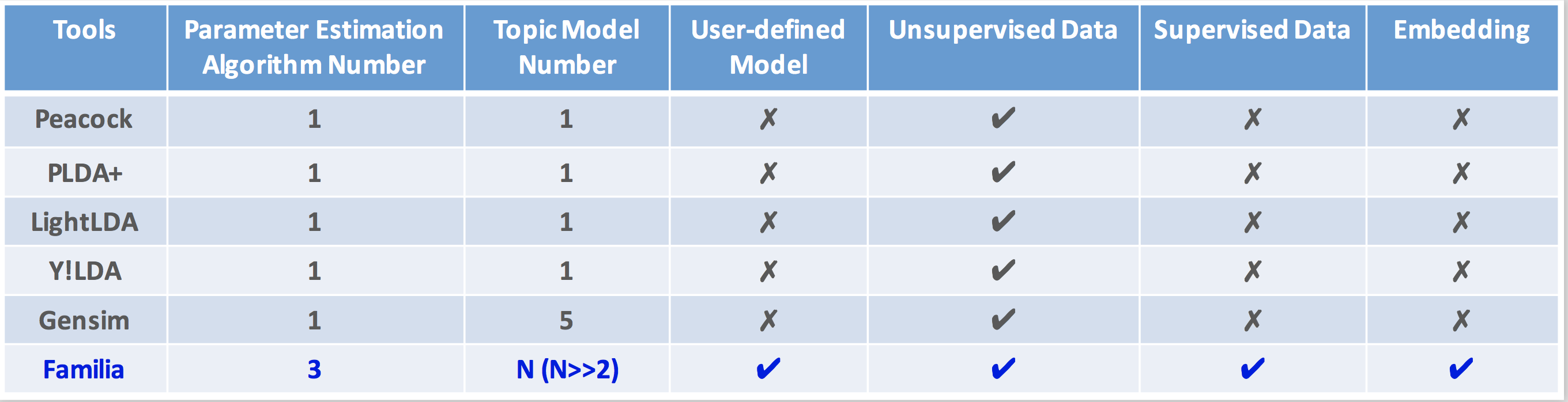}
\caption{Comparison between Familia and other topic modeling tools}
\label{pic:familia_comp}
\end{center}
\end{figure}

In this paper, we propose a novel topic modeling framework, Familia, which is easily configurable and can be utilized as off-the-shelf tool for software engineers without much knowledge of Bayesian networks. Compared with other industrial topic modeling tools (Fig.~\ref{pic:familia_comp}), Familia supports a broad line of topic models, which are of significant presence in the literature as well as heavily demanded in industry. Software engineers can investigate many topic models for their tasks at hand by simply changing the training data organization. Moreover, Familia takes over all the burdens of parameter inference, parallel computing and post-modeling utilities. Specifically, Familia contains three parameter inference methods: Gibbs sampling (GS) and Metropolis Hastings (MH) with alias table \cite{li2014reducing}, based on which a hybrid sampling mechanism is also designed to strike a balance between effectiveness and efficiency. To meet the requirements of topic modeling for massive data, Familia is inherently built upon the Parameter Server (PS) architecture \cite{dean2012large}\cite{xing2015petuum}\cite{li2013parameter}. Multiple computing nodes can be harnessed for parallel parameter inference and distributed storage. Furthermore, Familia contains multiple built-in post-modeling utilities such as dimensionality reduction and semantic matching, which can be readily applied in many downstream applications.

The contributions of this paper are summarized as follows:
\begin{enumerate}
\item We identify the huge gap between topic model research in academia and industrial practice of topic modeling. We propose a novel framework, named Familia, which bridges this gap. To the best of our knowledge, it is the first framework that supports multiple topic models in a user-friendly manner. A variety of topic models can be modeled by Familia, to name a few, LDA~\cite{blei2003latent}, Supervised LDA~\cite{mcauliffe2008supervised}, Sentence LDA~\cite{jo2011aspect}, TOT~\cite{wang2006topics}, Bilingual Topic Model~\cite{gao2011clickthrough}, Pair Model~\cite{jagarlamudi2013modeling}, GeoFolk~\cite{sizov2010geofolk}, LATM~\cite{wang2007mining} and Multifaceted Topic Model~\cite{vosecky2014integrating}. Besides these existing models, users can design their own topic models for specific tasks as long as they follow the generative process of Familia. 

\item We systematically investigate the performance of GS and MH with different topic models. We further propose a hybrid sampling mechanism to balance the effectiveness and efficiency of parameter inference. Based on the earned insights, we provide practical suggestions on choosing sampling method for different topic models.

\item 
We demonstrate the utilities and necessity of Familia in several real-life industrial cases and application that impact millions of users, which also acts like a guide for users to select appropriate topic models for their tasks and apply them in a proper way using Familia. Currently Familia has been widely used in both industrial community and academia, and has become the second most popular project (with more than 1.3k stars and 350 forks) among all the open-sourced topic model tools in Github. 
\end{enumerate}

The rest of this paper is organized as follows. In Section~\ref{sec:Related Work}, we review the related work. In Section~\ref{sec:Industrial Utilities}, we describe the utilities of topic models. 
In Section~\ref{sec:Familia}, we discuss the mathematical foundations of Familia. In Section~\ref{sec:parameter inference}, we detail the parameter inference algorithms. In Section~\ref{sec: System Implementation}, we describe some issues of system implementation. 
In Section~\ref{sec:experiments}, we present the experimental results. Then we discuss several industrial cases in Sections~\ref{sec:Industrial Cases}. Finally, we conclude this paper in Section~\ref{sec:Conclusion}.

\section{Related Work}
\label{sec:Related Work}

The present work is related to a wide range of topic models in the literature and the recent advances of parameter inference algorithms of LDA.

\subsection{Topic Models}

LDA \cite{blei2003latent} plays an important role in the field of topic modeling. In the last decade, many extensions of LDA have been proposed to meet specific needs of many applications. For example, \cite{wang2006topics} presented the Topic-over-Time (TOT) that captures latent topics and their changes over time.  Supervised LDA \cite{mcauliffe2008supervised} captured the regularity of labelled documents by introducing response signals. 
Location Aware Topic Model (LATM) \cite{wang2007mining} is designed to explicitly model the information between locations and words. 
GeoFolk \cite{sizov2010geofolk} focuses on discovering latent topics from social media by using text features as well as spatial information. 
\cite{jo2011aspect} proposed Sentence LDA that assumes all words from one sentence are generated from one topic. 
A bilingual topic model was proposed in \cite{gao2011clickthrough} as a language modeling framework, in which the topics are learned from query-title pairs. 
Multi-Faceted Topic Model (MfTM) \cite{vosecky2014integrating} was proposed to capture the temporal characteristics of each topic by jointly modeling latent semantics among terms and entities. Although the aforementioned ones take only a small portion of topic models in the literature, they are suitable choices in many industrial scenarios. 
However, to the best of our knowledge, none of them are supported by existing industrial topic model systems such as PLDA+ \cite{liu2011plda+}, Peacock \cite{wang2014towards} and Gensim \cite{rehurek_lrec}. However, compared to Familia, these existing systems only support limited kinds of topic models, and to the best of our knowledge, none of above topic models except LDA are supported by existing systems.

\subsection{Advanced Parameter Inference of LDA}
The latest advances of parameter inference of LDA roughly fall into two major categories: (1) Parallel parameter inference; (2) Efficient sampling algorithms. We review the related works from the two categories respectively.

\subsubsection{Parallel Parameter Inference}

Parallelization is a straightforward approach for speeding up  parameter inference. \cite{newman2009distributed} proposed distributed algorithm for LDA. Their algorithm partitioned the data across separate processors and conduct inference in a parallel approach. Each processor performs Gibbs sampling over local data and then a global synchronization merges updates from all processors. It was shown that the converged probability for distributed learning is similar to that obtained with single-processor.  Thereafter, \cite{wang2014towards} proposed a distributed topic modeling system named Peacock, which is a hierarchical distributed architecture for training LDA. More recently, \cite{yuan2015lightlda} proposed a model-parallel scheme that leverages dependency to efficiently train LDA and the scheme is frugal on memory consumption and network communication. 

\subsubsection{Efficient Sampling Algorithms}

Another major research trend of LDA parameter inference is to design efficient sampling algorithms. 
Conventional Gibbs sampling \cite{griffiths2004finding} has a complexity of $K$ per sample when the topic amount is set to $K$. 
\cite{porteous2008fast} introduced FastLDA, which has a complexity of significantly less than $K$ per sample.  
\cite{yao2009efficient} proposed SparseLDA that contains both an algorithm and data structures for efficiently conducting Gibbs sampling. 
\cite{li2014reducing} proposed an algorithm which achieves $O(K_d)$ sampling complexity, where $K_d$ is the instantiated topics in the document. 
As one step further, \cite{yuan2015lightlda} proposed LightLDA, which uses an Metropolis-Hastings sampling algorithm and achieves $O(1)$ time complexity per token. Recently, \cite{chen2016warplda}  proposed WarpLDA, which achieves both $O(1)$ time complexity per token and $O(K)$ scope of random access. With their effectiveness, these techniques are exclusively designed for LDA.  Their applicability has never been explored in a broader scope of topic models, on which how to conduct efficient sampling is still an open question.

\subsubsection{Variational Bayesian Inference}

Variational inference has also been widely used for approximating intractable integrals arising in Bayesian inference. 
The concept is initially from statistical physics, and \cite{Peterson1987a} adopts it to probabilistic inference, which fits a neural network using Mean-field variational inference methods. Meanwhile, \cite{Hinton1993Keeping} also combines variational algorithms with neural network models. 
In early 1990s, variational inference methods are generalized to many probabilistic models \cite{Saul1995Exploiting} \cite{Jaakkola1996A} and \cite{Jordan1999An}. 
LDA \cite{blei2003latent} relies on Mean-field variational inference method before various sampling algorithms becoming popular. 
Recent works in this area focus on making variational inference methods more scalable, easy to derive and supporting more complex models\cite{Kingma2013Auto} \cite{Ranganath2013Black}. For example, stochastic variational inference \cite{Hoffman2013Stochastic} scales variational inference to massive data. 

\section{Industrial Utilities}
\label{sec:Industrial Utilities}

Based on a comprehensive survey of industrial applications, we find that the industrial utilities of topic models essentially fall into two major categories: (1) dimensionality reduction; (2) semantic matching. In this section, we describe the details of these two utilities, which motivate the design philosophy of Familia.

\subsection{Dimensionality Reduction}

Topic models are powerful tools for representing high dimensional data in a lower dimension \cite{yao2009efficient}. Through applying topic models, each ``document'' can be represented with its topic distribution. This can be considered as dimensionality reduction, since the topic space is typically much smaller than the original data space. The topical representation of each document can be conveniently utilized as features in downstream tasks such as document clustering and document classification \cite{blei2003latent}.

\subsection{Semantic Matching}

Another utility of topic models is semantic matching, i.e., calculating a score to indicate the relevance between a ``query'' $q$ and a ``document'' $d$ \cite{wei2006lda}\cite{si2010confucius}.  There exist two paradigms of deriving the score. The first one is based upon Jensen-Shannon divergence \cite{lin1991divergence}:

\begin{equation}
\begin{split}
Score(q,d)&=1-{{\rm {JSD}}}(t_q\parallel t_d)=1-{\frac  {1}{2}}D(t_q\parallel M)-{\frac  {1}{2}}D(t_d\parallel M),
\end{split}
\end{equation}
where $t_q$ is the topic distribution of $q$, $t_d$ is the topic distribution of $d$, ${M={\frac {1}{2}}(t_q+t_d)}$ and $D(\cdot)$ stands for Kullback-Leibler divergence. A more convenient approach is based on calculating the overall probability of each item in $q$ generated by $d$:
\begin{equation}
\label{alg:semantic2}
\begin{aligned}
Score(q,d)=\prod_{w\in q}P(w|d)=\prod_{w\in q}\sum_{z}P(w|z)P(z|d),
\end{aligned}
\end{equation}

where $w$ is the item in $q$, $P(z|d)$ is the probability of $d$ generating $z$ and $P(w|z)$ is the probability of $z$ generating $w$.

\section{Generative Assumption}
\label{sec:Familia}

In this section, we discuss the mathematical foundations of Familia. Since Familia aims to define the generative process of a series of topic models, we abstract the data organization and structure shared by all these models. We describe Familia in conventional topic model terminologies, and three newly introduced terminologies are formally defined as follows:

\begin{enumerate}
\item \it{blob}: the basic unit in which all the items are generated by the same topic, such as a sentence for SentenceLDA. 

\item \it{factor}: the basic unit in which all the item are generated by the same distribution and the same topic; for example, all the words in one sentence usually comprise one factor for TOT. In terms of the distribution being continuous or discrete, the factors can be further categorized as continuous factor and discrete factor. For example, all the timestamps in one sentence usually belong to the same continuous factor for TOT. 

\item \it{item}: the basic unit of an observed variable, such as a word, a timestamp, etc
\end{enumerate}

\begin{algorithm}
\caption{Generative Process of Familia}
\label{alg:Generative Process of Familia}
\For {each topic $k \in {1,...,K}$}
{
\For {each discrete factor $i \in {1,...,M}$}
{
draw a discrete factor distribution $\phi_{ik} \sim$ Dirichlet$(\beta_i)$
}
\For {each continuous factor $j \in {1,...,N}$}
{
 generate a continuous factor distribution $\psi_{jk}$
 }
}
\For {each document $d \in {1,...,D}$}
{
generate topic distribution $\theta_{d} \sim$ Dirichlet$(\alpha)$\\
\For {each blob $b$ in $d$}
{
generate a topic $z \sim \theta_{d}$ \\
\For {each discrete factor $i \in {1,...,M}$}
{
 generate items $u \sim \phi_{iz}$
}
\If{there exists continuous factor}
{\For {each continuous factor $j \in {1,...,N}$}
{
 generate items $v \sim \psi_{jz}$
 }
 }
}
\If{ there exists supervised signal}
{
draw signal $y_d \sim P(y_d|\overline{z}_d, \eta, \sigma^2)$, where $P(y_d|\overline{z}_d, \eta, \sigma^2) = N(y_d|\eta^T\overline{z}_d, \sigma^2)$
}
}
\end{algorithm}

The generative process is depicted in Algorithm~\ref{alg:Generative Process of Familia}. In order to generate a document $d$, we first draw  $\theta_{d}$, which is a Multinomial distribution over topics. Then, for each blob, we draw a topic $z$.  Based upon $z$, for each discrete factor $i$, we generate discrete items $u$ according to the corresponding discrete distribution $\phi_{iz}$. The continuous items are generated in an analogous approach. Finally, if the topic model needs to capture the supervised signal (e.g., the category of the document or the rating of the quality of the document), the signal is further drawn from a Gaussian distribution that uses $\overline{z}_d$ as parameters.  Specifically, $\overline{z}_d=\frac{1}{N}\sum_{n=1}^N z_n$, where $N$ is the amount of tokens in the document. 

It is easy to see that Algorithm~\ref{alg:Generative Process of Familia} is a generic process for many topic models.  A variety of topic models can be modeled by Algorithm~\ref{alg:Generative Process of Familia} , to name a few, LDA\cite{blei2003latent} , Supervised LDA\cite{mcauliffe2008supervised}, Sentence LDA\cite{jo2011aspect}, TOT\cite{wang2006topics}, Bilingual Topic Model\cite{gao2011clickthrough}, Pair Model\cite{jagarlamudi2013modeling}, GeoFolk \cite{sizov2010geofolk}, LATM\cite{wang2007mining}  and Multifaceted Topic Model\cite{vosecky2014integrating} . Besides these existing models, users can design their own topic models for specific tasks as long as they follow the generative process in Algorithm~\ref{alg:Generative Process of Familia}. It is not difficult to see that the topic models supported by Familia can be conveniently applied to scenarios where the two utilities discussed in Section~\ref{sec:Industrial Utilities} are used.

\section{Parameter Inference}
\label{sec:parameter inference}

We proceed to discuss how to conduct parameter inference for Familia. We detail the mathematical derivation for the most complicated scenario where there simultaneously exist discrete factors, continuous factors and the supervised response. The other simpler scenarios can be trivially derived based upon the following discussion. The generative process of Algorithm~\ref{alg:Generative Process of Familia} can be By translated into joint distribution, and the objective is to maximize the likelihood of the observed items and the supervised signals. The complete likelihood $P(\mathbf{u}_{1...M}, \mathbf{v}_{1...N}, \mathbf{y}_{1...D} , \mathbf{z}|\alpha, \beta, \Psi, \eta, \sigma)$ is presented as follows:

\begin{equation}
\label{eq:prob}
\small
\begin{aligned}
P(\mathbf{u}_{1...M}, \mathbf{v}_{1...N} ,\mathbf{y}_{1...D} , \mathbf{z}|\alpha, \beta, \Psi, \eta, \sigma)=\\
P(\mathbf{z}|\alpha)\underbrace{\prod_{i=1}^MP(\mathbf{u}_{i} |\mathbf{z}, \beta)}_{discrete\quad factors}
\underbrace{\prod_{j=1}^NP( \mathbf{v}_{j}|\mathbf{z}, \Psi)}_{continuous\quad factors}
\underbrace{\prod_{d=1}^DP(y_d|\overline{z}_d, \eta, \sigma^2)}_{supervised\quad signal}\\
=\Big(\frac{\Gamma(\sum_{z=1}^T\alpha_z)}{\prod_{z=1}^T\Gamma(\alpha_z)}\Big)^D\prod_{d=1}^D\frac{\prod_{z=1}^T\Gamma(m_{dz}+\alpha_z)}{\Gamma(\sum_{z=1}^T(m_{dz}+\alpha_z))}\\
\prod_{i=i}^M\Big(\frac{\Gamma(\sum_{u_i=1}^{U_i}\beta_{u_i})}{\prod_{u_i=1}^{U_i}\Gamma(\beta_{u_i})}\Big)^T  \prod_{z=1}^{T}\frac{\prod_{u_i=1}^{U_i}\Gamma(n_{z{u_i}}+\beta_{u_{i}})}{\Gamma(\sum_{u=1}^{U_i}(n_{zu_i}+\beta_{u_i}))}\\
\prod_{j=1}^N\prod_{d=1}^D\prod_{l=1}^{L_d}P(v_{jdl} |\psi_{z_{jdl}})
\prod_{d=1}^DP(y_d|\overline{z}_d, \eta, \sigma^2)
\end{aligned}
\end{equation}
where $m_{dz}$ denotes the number of sentences that are generated by topic $z$ in document $d$. $n_{zv}$ is the number of times that $v$ is generated by topic $z$ through Multinomial distribution and $\Gamma(\cdot)$ indicates Gamma function. The goal of parameter inference is to estimate $\Theta$, $\Phi$, $\Psi$, $\eta$ and $\sigma$ in Algorithm~\ref{alg:Generative Process of Familia} through sampling the latent topic $z$ of each blob. In this following two subsections, we describe how to sample $\mathbf{z}$ through Gibbs sampling and Metropolis Hastings based on Eq.~(\ref{eq:prob}). Utilizing the sampled $z$ to estimate the parameters of discrete distributions is well-documented in literature \cite{jo2011aspect}. We will detail how to utilize $\mathbf{z}$ to estimate the parameters of continuous distribution in Section~\ref{sec:Issues of Continuous Distributions}.

\subsection{Gibbs Sampling}

According to Bayes rule and Eq.~(\ref{eq:prob}), the full conditional of blob $b$ belonging to topic $k$ in document $d$ is as follows:

\begin{equation}
\label{eq:GS sampling}
\small
\begin{aligned}
P(z_{b}=k |\mathbf{u}_{1...M}, \mathbf{v}_{1...N}, \mathbf{y}_{1...D}, \mathbf{z}_{-b}, \alpha, \beta, \Psi, \eta, \sigma)= \\
\frac{m_{dk}+\alpha_k}{\sum^K_{k'=1}(m_{dk'}+\alpha_{k'})} \\
\prod_{i=i}^M\frac{\Gamma(\sum^{U_i}_{u=1}(n_{ku}+\beta_{u}))}{\Gamma(\sum^{U_i}_{u=1}(n_{ku}+\beta_{u}+N_{bu}))}
\prod_{u_i\in b}\frac{\Gamma(n_{ku_i}+\beta_{u_i}+N_{bu_i})}{\Gamma(n_{ku_i}+\beta_{u_i})} \\
\prod_{j=1}^N\prod_{l=1}^{L_b}P(v_{jl} | z_k)\times
P(y_d|\overline{z}_d, \eta, \sigma^2)
\end{aligned}
\end{equation}

\noindent In order to sample a new topic for the blob $b$, we need to calculate the above conditional probability for all topics and conduct normalization. Hence, the time complexity for sampling a topic for a blob is $O(K)$, where $K$ is the topic amount.

\subsection{Metropolis Hastings}

If the topic amount $K$ is large, the $O(K)$ per blob complexity is time-consuming. We now propose an efficient alternative based upon Metropolis Hastings (MH), which has been successfully applied in \cite{li2014reducing} \cite{yuan2015lightlda} for LDA. For the convenience of designing proper proposals for MH, we first conduct approximation to Eq.~(\ref{eq:GS sampling}) as follows:

\begin{equation}
\label{eq:MH_sampling}
\begin{aligned}
Q(z_{b}=k |\mathbf{u}_{1...M}, \mathbf{v}_{1...N}, \mathbf{y}_{1...D}, \mathbf{z}_{-b}, \alpha, \beta, \Psi, \eta, \sigma) \approx \\
\underset{\textup{document-topic proposal}}{\underbrace{\frac{m_{dk}+\alpha_k}{\sum^K_{k'=1}(m_{dk'}+\alpha_{k'})}}} \\
\prod_{i=i}^M\prod_{u_i\in b} \underset{\textup{discrete item-topic proposal}}{\underbrace{\frac{n_{ku_i}+\beta_{u_i}}{\sum^{U_i}_{u=1}(n_{ku}+\beta_{u})}}}
\prod_{j=1}^N\prod_{l=1}^{L_b} \underset{\textup{continuous Item-topic proposal}}{\underbrace{P(v_{jl} | z_k)}}
\times
\underset{\textup{supervised signal proposal}}{\underbrace{ P(y_d|\overline{z}_d, \eta, \sigma^2)}}
\end{aligned}
\end{equation}

\noindent The above equation approximates Eq.~(\ref{eq:GS sampling}) when an item appears multiple times with in a blob. Based on this approximation, we discuss the MH sampling method for Familia. MH needs proper proposals to work.  We now discuss four proposals which fall into two major categories: document-based proposal and item-based proposal.

$\bullet$ Document-based Proposals:

The first document-specific proposal is the \textit{document-topic proposal}:

\begin{equation}
\label{equ:doc-top}
Q_{d}(k) \propto \frac{m_{dk}+\alpha_k}{\sum^K_{k'=1}(m_{dk'}+\alpha_{k'})}
\end{equation}

According to the MH algorithm, the acceptance ratio from state $i$ to $j$ is: 

\begin{equation}
\label{equ:accept-doc-top}
\textup{min }\{1, \frac{Q(j)Q_d(i)}{Q(i)Q_d(j)}\}
\end{equation}

The second document-specific proposal is the \textit{supervised signal proposal}:

\begin{equation}
\label{equ:sup-sig}
Q_{y_d}(k) \propto P(y_d|\overline{z}_d, \eta, \sigma^2)
\end{equation}

According to the MH algorithm, the acceptance ratio from state $i$ to $j$ is: 

\begin{equation}
\label{equ:accept-sup-sig}
\textup{min }\{1, \frac{Q(j)Q_{y_d}(i)}{Q(i)Q_{y_d}(j)}\}
\end{equation}

$\bullet$ Item-based Proposals:

The first item-based proposal is the \textit{discrete item-topic proposal}, which is denoted as $Q_{u}(k)$:

\begin{equation}
\label{equ:item-top}
Q_{u}(k) \propto \frac{n_{ku_i}+\beta_{u_i}}{\sum^{U_i}_{u=1}(n_{ku}+\beta_{u})}
\end{equation}

According to the MH algorithm, the acceptance ratio from state $i$ to $j$ is: 

\begin{equation}
\label{equ:accept-item-top}
\textup{min }\{1, \frac{Q(j)Q_{u}(i)}{Q(i)Q_{u}(j)}\}
\end{equation}

The second item-based proposal is the \textit{continuous Item-topic proposal}, which is denoted as $Q_{v}(k)$:

\begin{equation}
\label{equ:item-top-con}
Q_{v}(k) \propto P(v_{jl} | z_k)
\end{equation}

According to the MH algorithm, the acceptance ratio from state $i$ to $j$ is: 

\begin{equation}
\label{equ:accept-item-top-con}
\textup{min }\{1, \frac{Q(j)Q_{v}(i)}{Q(i)Q_{v}(j)}\}
\end{equation}

\begin{figure*}[ht!]
  \centering
  \includegraphics[width=0.75\textwidth]{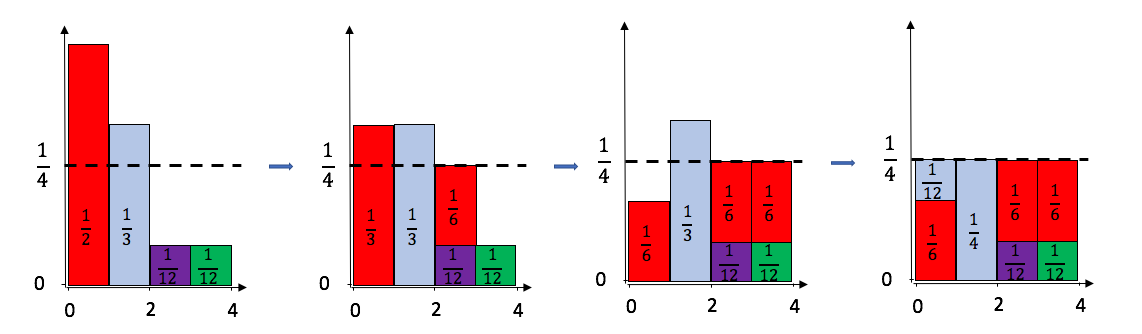}
  \caption{An example of Alias Table construction, which transform a non-uniform sampling process into a uniform sampling one. The whole process enables O(1) amortized sampling complexity.}
  \label{pic:alias}
\end{figure*}

It is easy to see that each proposal encourages the sparsity of their corresponding component in Eq.~(\ref{eq:MH_sampling}). For fast sampling from each proposal, a data structure named alias table \cite{li2014reducing}\cite{yuan2015lightlda} is utilized to reduce the sampling complexity (see Fig.~\ref{pic:alias}). Theoretically, an alias table need to be created for each document and each item. A caveat is that document-topic proposal does not need to explicitly establish alias table \cite{yuan2015lightlda}, because sampling from the document-topic proposal can be cheaply simulated through returning the topic assignment of a randomly sampled blob in the document. The MH method is formally presented in Algorithm~\ref{alg:Metropolis Hastings}. When sampling a new topic for a blob, the algorithm sequentially utilizes document-based proposals and item-based proposals to update the topic candidates.  For each topic candidate, the algorithm chooses whether to accept it according to the acceptance ratio, just like the standard Metropolis Hastings.  Note that this process can be repeated for several iterations and each iteration is formally defined as an \emph{MH step}. In the experiment section, we will show the extent to which the number of MH steps can affect the performance of the MH method.

\begin{algorithm}
\caption{Metropolis Hastings of Familia}
\label{alg:Metropolis Hastings}
\For {each document $d \in {1,...,D}$}
{
\For {each blob $b$ in $d$}
{
\For {a predefined number of MH steps}
{
propose a topic $z_1$ based on document-topic proposal according to Eq.~\eqref{equ:doc-top}\\
update the topic to $z^{'}_1$ according to acceptance ratio according to Eq.~\eqref{equ:accept-doc-top}\\
propose a topic $z_2$ based on alias table of $y_d$ according to Eq.~\eqref{equ:sup-sig}\\
update the topic to $z^{'}_2$ according to acceptance ratio according to Eq.~\eqref{equ:accept-sup-sig}\\
\For {each discrete factor $i \in {1,...,M}$}
{
\For{item $u$ in this factor}
{
propose a topic $z_{u3}$ based on alias table of $u$ according to Eq.~\eqref{equ:item-top}\\
update the topic to $z^{'}_{u3}$ according to acceptance ratio according to Eq.~\eqref{equ:accept-item-top}\\
}
}
\For {each continuous factor $j \in {1,...,N}$}
{
\For{item $v$ in this factor}
{
propose a topic $z_{v4}$ based on alias table of $v$ according to Eq.~\eqref{equ:item-top-con}\\
update the topic to $z^{'}_{v4}$ according to acceptance ratio according to Eq.~\eqref{equ:accept-item-top-con}\\
}
}
}
}
}
\end{algorithm}

\subsection{Issues of Continuous Distributions}
\label{sec:Issues of Continuous Distributions}

For topic models with continuous factors, updating the parameters of continuous distributions is computationally expensive, especially for distributed environment in which the synchronization of these parameters is needed. In Familia, we choose a well-adopted practice \cite{wang2006topics}\cite{sizov2010geofolk}:
we update the parameters of the continuous distributions after each major iteration (i.e., scanning through the whole corpus of training data). For Gaussian distributions, we straightforwardly update the parameters by the sample mean and sample variance. If the continuous distribution is Beta distribution $Beta({\psi}_{k1}, {\psi}_{k2})$, we update the parameters ${\psi}_{k1}$ and ${\psi}_{k2}$ for the $k$th topic as follows:

\begin{equation}
\small
\label{eq:t1}
{\psi}_{k1}=\bar{s_{k}}(\frac{\bar{s_k}(1-\bar{s_k})}{v^2_{k}}-1),
\end{equation}

\begin{equation}
\small
\label{eq:t2}
{\psi}_{k2}=(1-\bar{s_{k}})(\frac{\bar{s_k}(1-\bar{s_k})}{v^2_{k}}-1),
\end{equation}
where $\bar{s_k}$ and $v^2_k$ denote the sample mean and biased sample variance of topic $k$'s items. As for supervised signals \cite{mcauliffe2008supervised}, we denote the $(D\times K)$ matrix whose $d$th row is $\overline{z}^T_d$ as $A$,  and the $D\times 1$ vector of supervised signals as $Y$, the $\eta$ and $\sigma$ are updated as follows:

\begin{equation}
\small
\eta = (A^TA)^{-1}A^TY
\end{equation}

\begin{equation}
\small
\sigma = \frac{1}{D}(Y^TY-Y^TA(A^TA)^{-1}A^TY)
\end{equation}

\noindent In practice, the above matrix manipulation can be approximated by techniques to reduce the computational cost and scale to large data set.

\subsection{Hybrid Sampling Mechanism}

So far, we have discussed two basic sampling techniques that are supported in Familia.  The major advantage of MH is efficiency: the per blob sampling complexity can be reduced as low as $O(1)$ for some topic models through utilizing alias tables. However, as we will show later in Section~\ref{sec:experiments}, the models trained by GS are usually better than MH. Hence, it is desirable to design a mechanism that tradeoffs the efficiency of MH and effectiveness of GS. To meet this requirement, Familia enables the users to choose the sampling method for each iteration. Such flexibility makes it possible to investigate many hybrid sampling mechanisms which collectively apply GS and MH.

\section{System Implementation}
\label{sec: System Implementation}

In this section, we describe some important issues about system implementation of Familia. In Section~\ref{sec:Data Organization}, we describe the data organization of Familia. In Section~\ref{sec:Data and Parameter Storage}, we discuss the data and parameter storage.

\subsection{Data Organization}
\label{sec:Data Organization}

In Familia, data organization is critical for automating parameter inference because it provides basic information of Bayesian network structure of the topic model.  Based on data organization, Familia can deduce each component in Eq.~(\ref{eq:GS sampling}) and Eq.~(\ref{eq:MH_sampling}) and then conduct parameter inferences without imposing any burden on human to derive the mathematical equations. Documents are grouped as blocks to facilitate distributed computing. In each document, the blob is utilized as the basic unit whose content shares the same topic. A blob contains multiple factors and a factor can contain any amount of items. The data organization of Familia is presented as follows:

\begin{displaymath}
\small
  \text{block}
   \begin{cases}
    \text{document}_1
    \begin{cases}
      \text{supervised signal} \\
      \text{blob}_1
      \begin{cases}
      \text{factor}_1
      \begin{cases}
      \text{item}_1 \\
      \text{item}_2 \\
      \text{...}
      \end{cases}
      \\
      \text{factor}_2 \\
      \text{......}
      \end{cases}
      \\
      \text{blob}_2\\
      \text{......}
    \end{cases}
    \\
    \text{document}_2\\
    \text{......}
  \end{cases}
\end{displaymath}

\noindent The above data organization is generic. In practice, the ``item'' in Familia data organization can be specialized into words(discrete), tags(discrete) or even timestamps(continuous). 

\begin{figure*}[ht!]
\centering
\subfigure[LDA]
{
\label{fig:lda}
\includegraphics[width=0.2\textwidth]{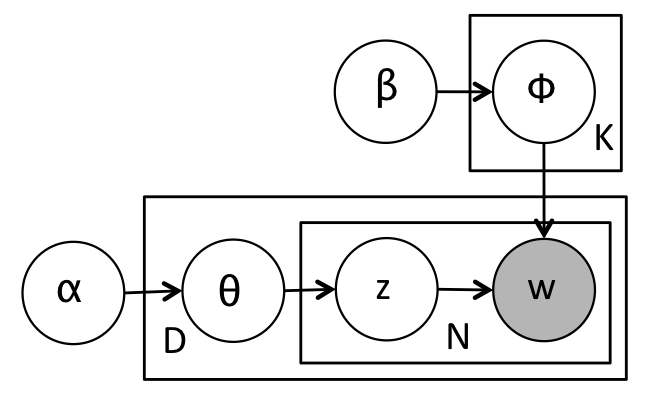}
}
{
\includegraphics[width=0.25\textwidth]{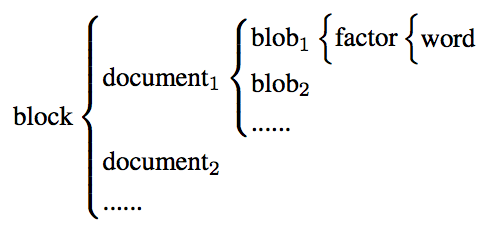}
}
\subfigure[Sentence LDA]
{
\label{fig:slda}
\includegraphics[width=0.2\textwidth]{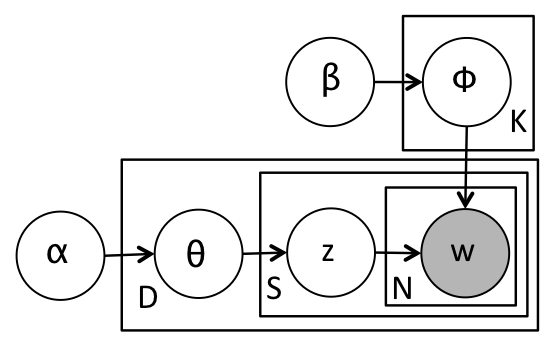}
}
{
\includegraphics[width=0.25\textwidth]{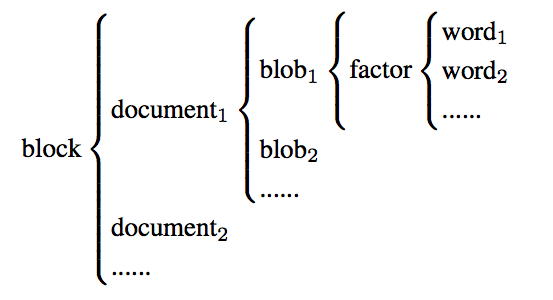}
}
\quad
\subfigure[Supervised LDA]
{
\label{fig:suplda}
 \includegraphics[width=0.2\textwidth]{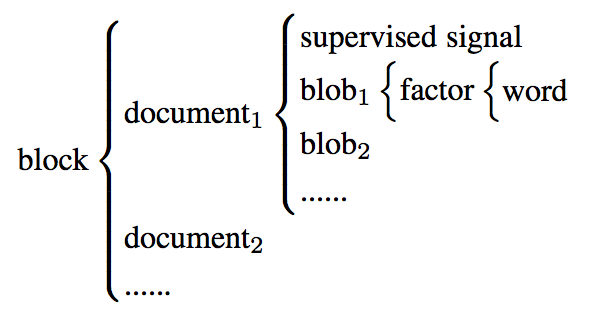}
}
{
 \includegraphics[width=0.25\textwidth]{figure3c2.png}
}
\subfigure[TOT]
{
\label{fig:tot}
\includegraphics[width=0.2\textwidth]{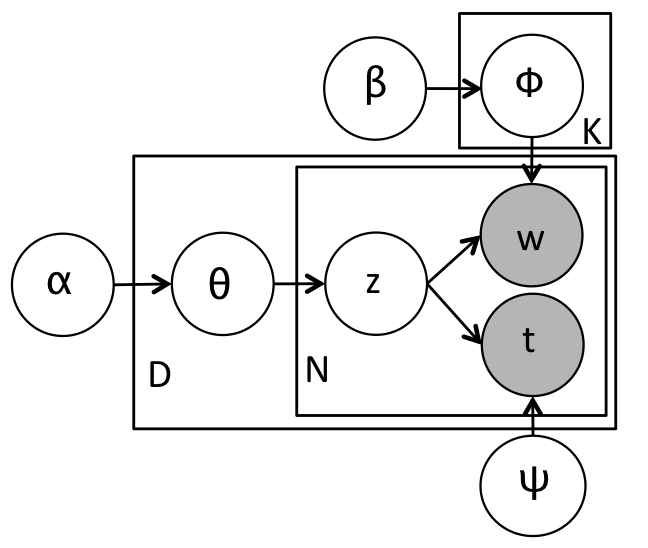}
}
{
\includegraphics[width=0.25\textwidth]{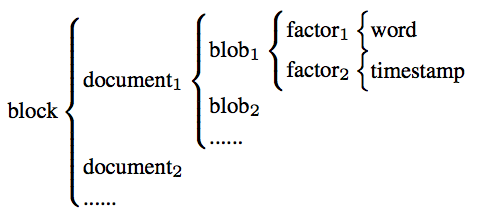}
}
\caption{Common Topic Models Examples and their Corresponding Data Organization in Familia}
\label{fig:Topic Models in Familia}
\end{figure*}

\noindent Now we demonstrate how to specialize the above data organization for several common topic models in Figure~\ref{fig:Topic Models in Familia}. For example, the data organization of LDA is described in Figure~\ref{fig:lda}. 

LDA only contains one kind of items: the words. It assumes that words are exchangeable in document. Hence, each blob only contains one factor and each factor contains one single word.  The data organization of TOT is described in Figure~\ref{fig:tot}. 

Each blob in TOT data organization contains two kinds of factors. Relying on such data organization, it is easy to see that GS and MH discussed in Section~\ref{sec:parameter inference} can be automatically conducted for topic models which are described by Algorithm~\ref{alg:Generative Process of Familia}. 

\subsection{Data and Parameter Storage}
\label{sec:Data and Parameter Storage}

    \begin{figure}[htb]
            \begin{center}
                \includegraphics[width=0.5\textwidth]{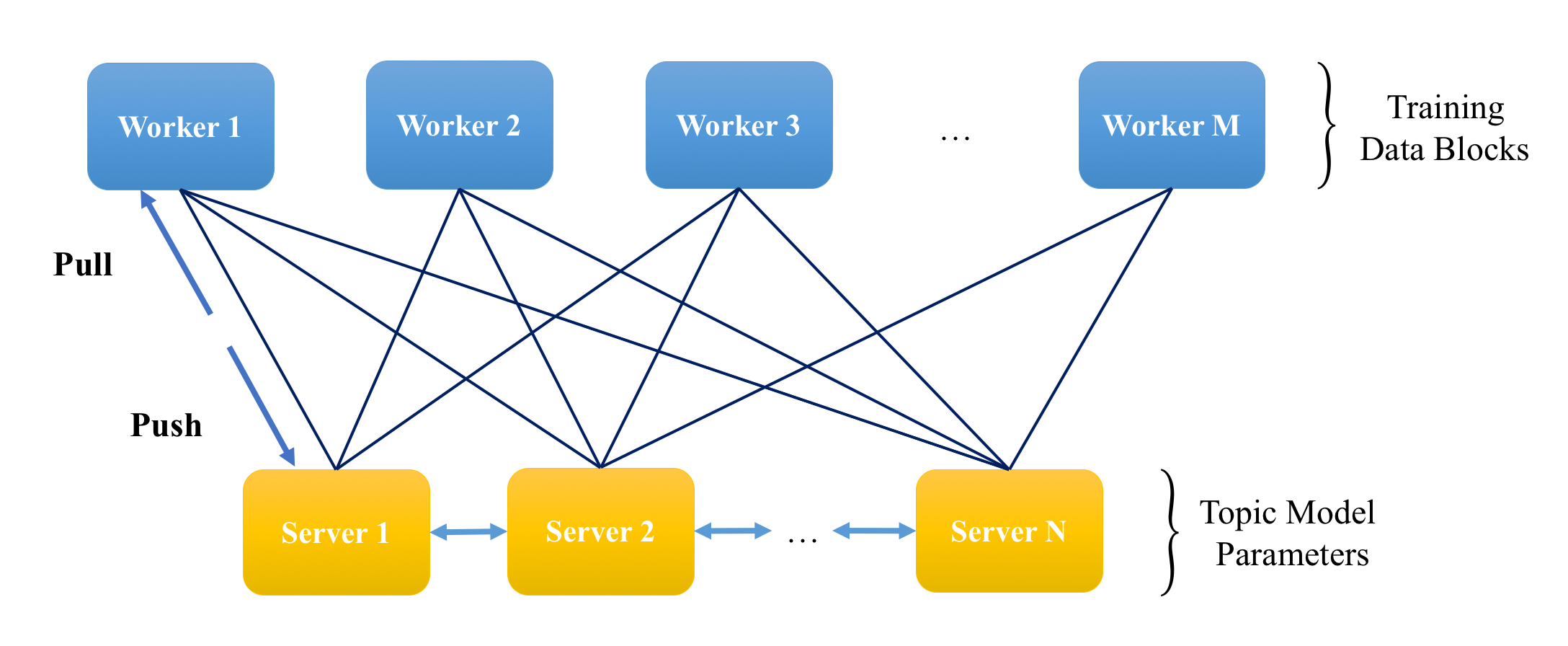}
                \caption{Parameter Server Architecture}
                \label{pic:ps}
            \end{center}
    \end{figure}

Familia adopts classic Parameter Server (PS) \cite{xing2015petuum}\cite{li2013parameter} architecture (see Figure~\ref{pic:ps})  for distributed computing. Both training data and topic model parameters are distributed on multiple computing nodes. The training data (i.e., the training data blocks) are sharded by their identifiers on workers while topic model parameters of discrete distributions are sharded by item identifiers on servers.  Each worker only pulls the parameters that are needed for processing the current block to the local memory. After the sampling procedure, each worker pushes the update information back to servers.  Detailed description of PS is beyond the scope of this paper, interested readers may refer to \cite{xing2015petuum}\cite{li2013parameter} for more information.

\section{Experiments}
\label{sec:experiments}

In this section, we report the experimental results. We first investigate the performance of a series of sampling methods across topic models and data sets in Section~\ref{sec:Performance of Sampling Methods}. Then we demonstrate the scalability of Familia on multiple computing nodes in Section~\ref{sec:Scalability of Familia}. 

\begin{figure*}
\centering
\subfigure[LDA (K=50)]
{
\includegraphics[width=0.3\textwidth]{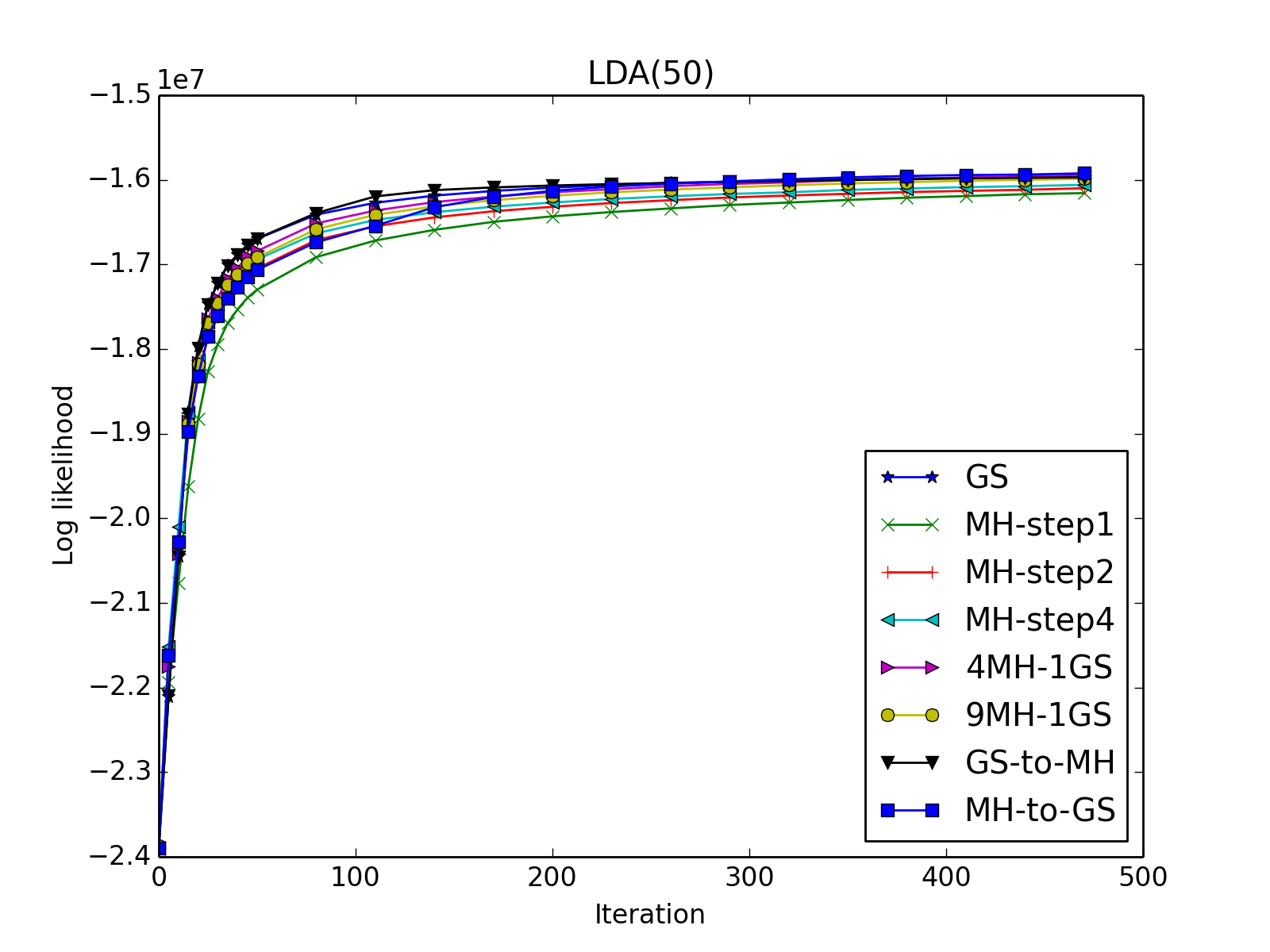} 
}
\subfigure[LDA (K=100)]
{
\includegraphics[width=0.3\textwidth]{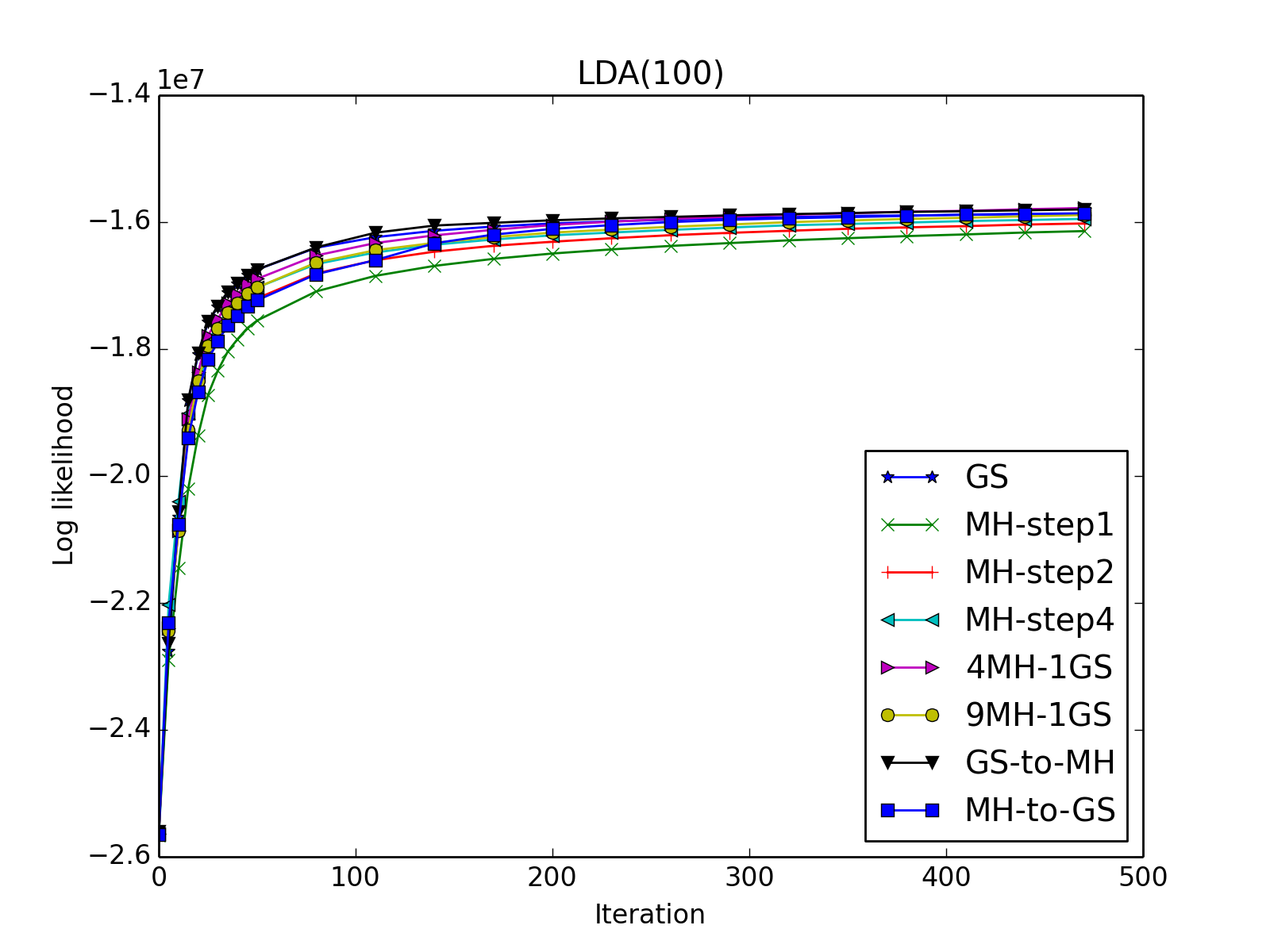} 
}
\quad
\subfigure[Sentence LDA (K=50)]
{
 \includegraphics[width=0.3\textwidth]{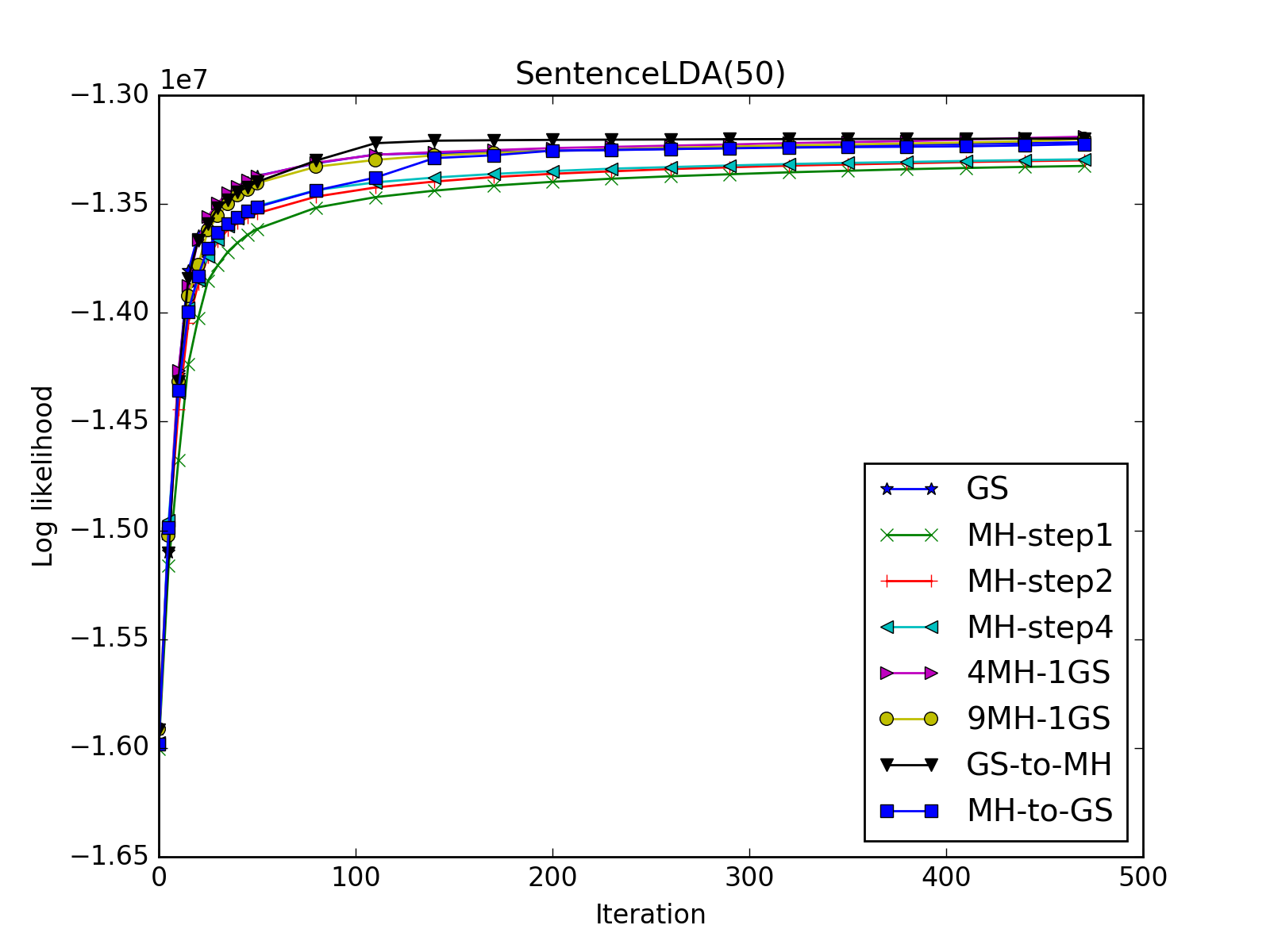}  
}
\subfigure[Sentence LDA (K=100)]
{
\includegraphics[width=0.3\textwidth]{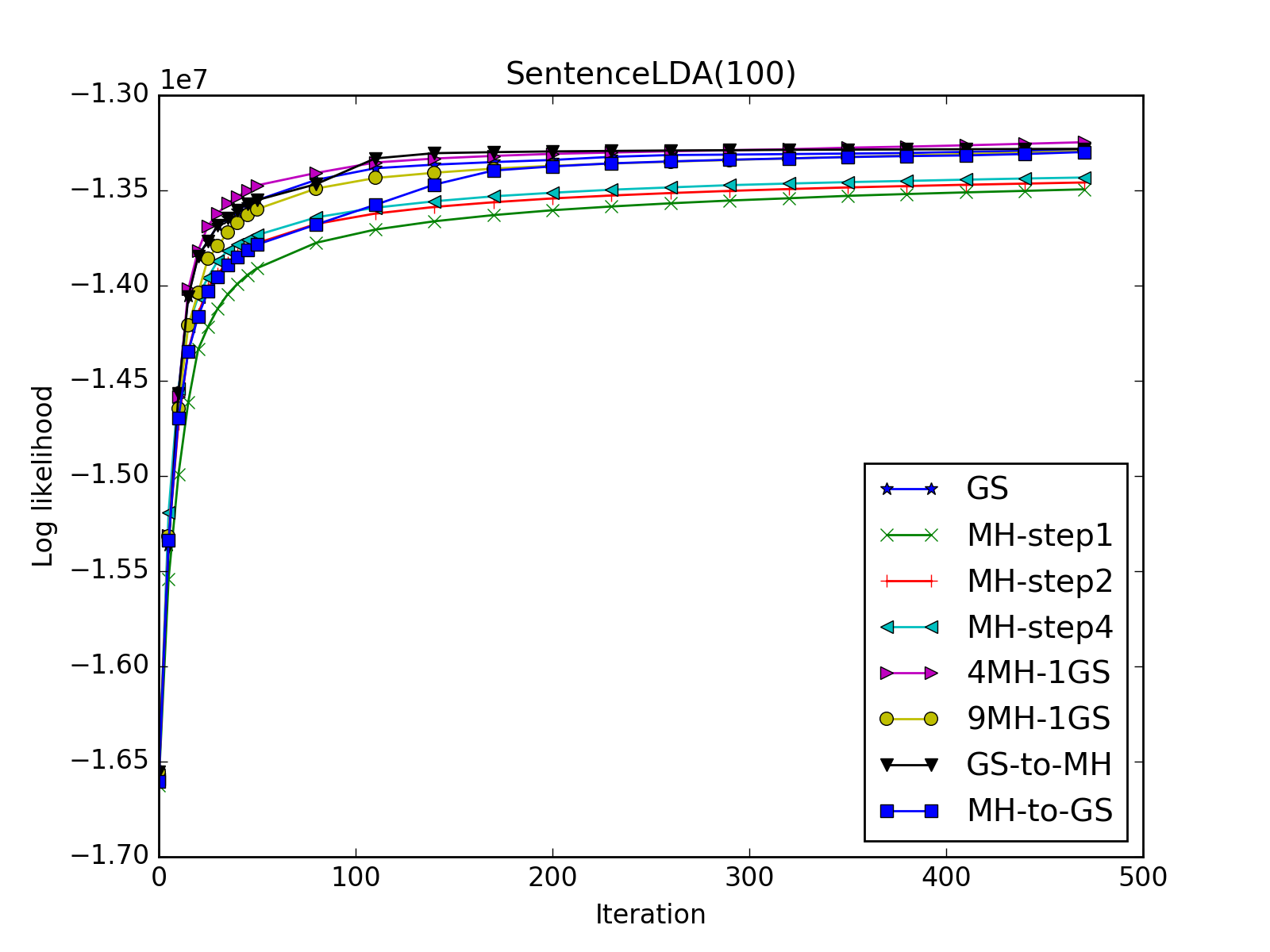}  
}
\quad
\subfigure[TOT (K=50)]
{
 \includegraphics[width=0.3\textwidth]{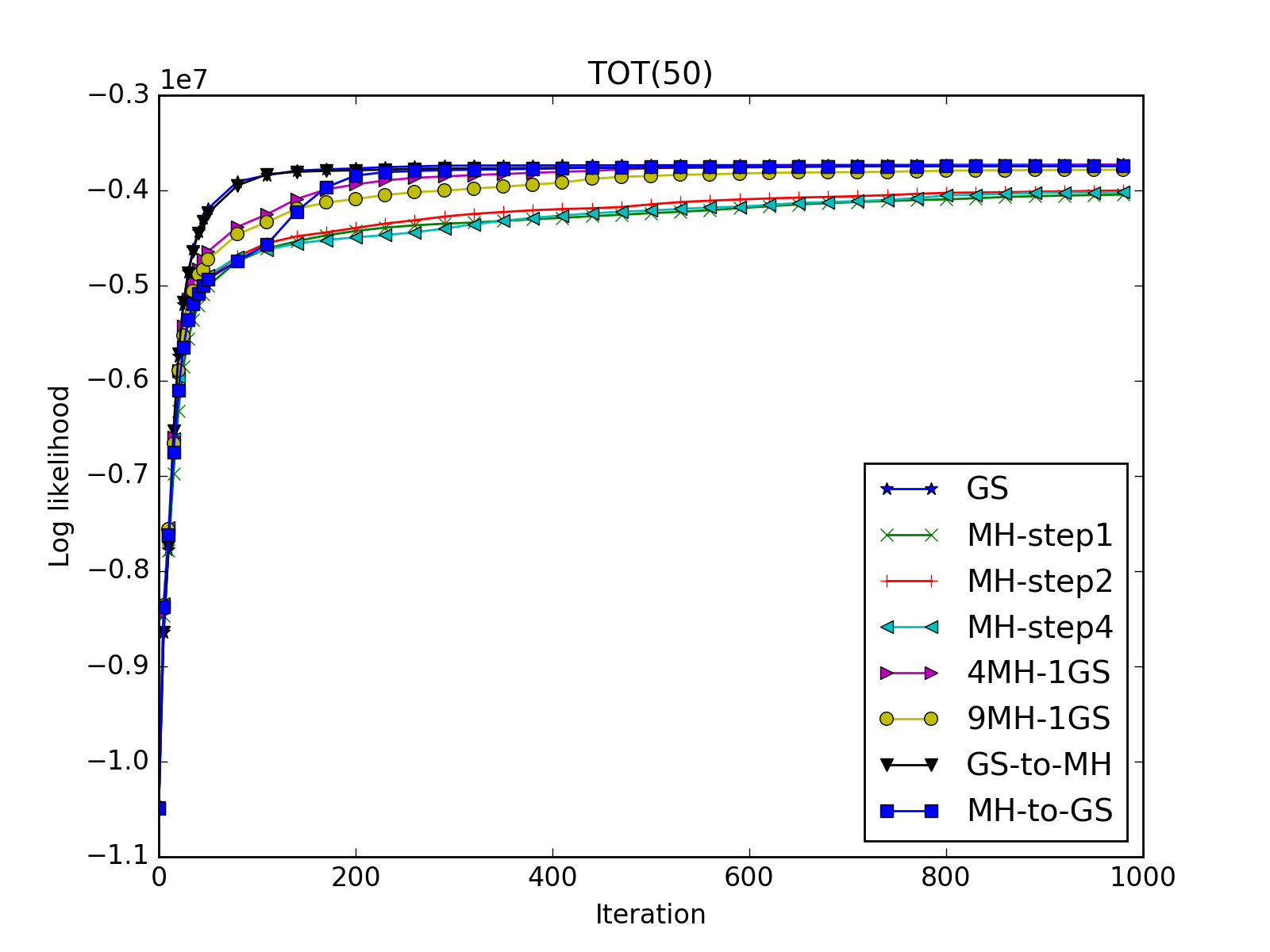} 
}
\subfigure[TOT (K=100)]
{
 \includegraphics[width=0.3\textwidth]{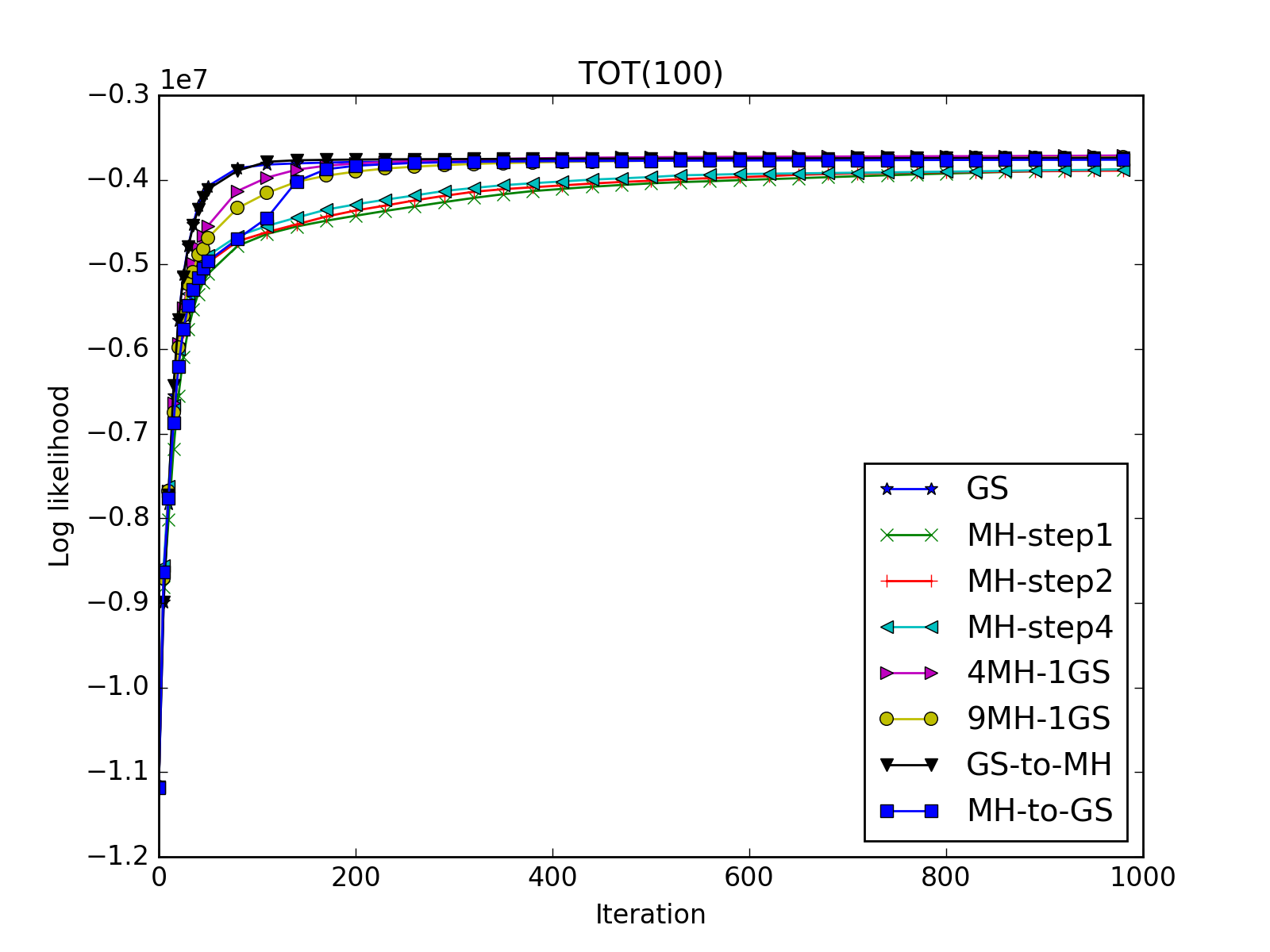}  
}
\caption{Comparison of Sampling Methods (Iteration) (Best Viewed in Color)}
\label{fig:Comparison of Sampling Methods (Iteration)}
\end{figure*}

\subsection{Performance of Sampling Methods}
\label{sec:Performance of Sampling Methods}

We systematically investigate the performance of different sampling methods in terms of LDA, Sentence LDA and TOT. In order to minimize effect beyond algorithmic performance, experiments in this subsection are conducted on a single computing node. NIPS dataset from UCI Bag of Words Data Set\footnote{https://archive.ics.uci.edu/ml/datasets/Bag+of+Words} is utilized for experiments of LDA.  Amazon data\footnote{http://uilab.kaist.ac.kr/research/WSDM11} is utilized for Sentence LDA. As for TOT,  we utilize the 21 decades of U.S. Presidential State-of-the-Union Addresses\footnote{http://www.gutenberg.org/dirs/etext04/suall11.txt} for the experiments. Due to space limitation, we present the results when the topic amount is set to 50 and 100. Similar insights can be obtained when the topic amount is set to other values.

\begin{figure*}
\centering
\subfigure[LDA (K=50)]
{
\includegraphics[width=0.3\textwidth]{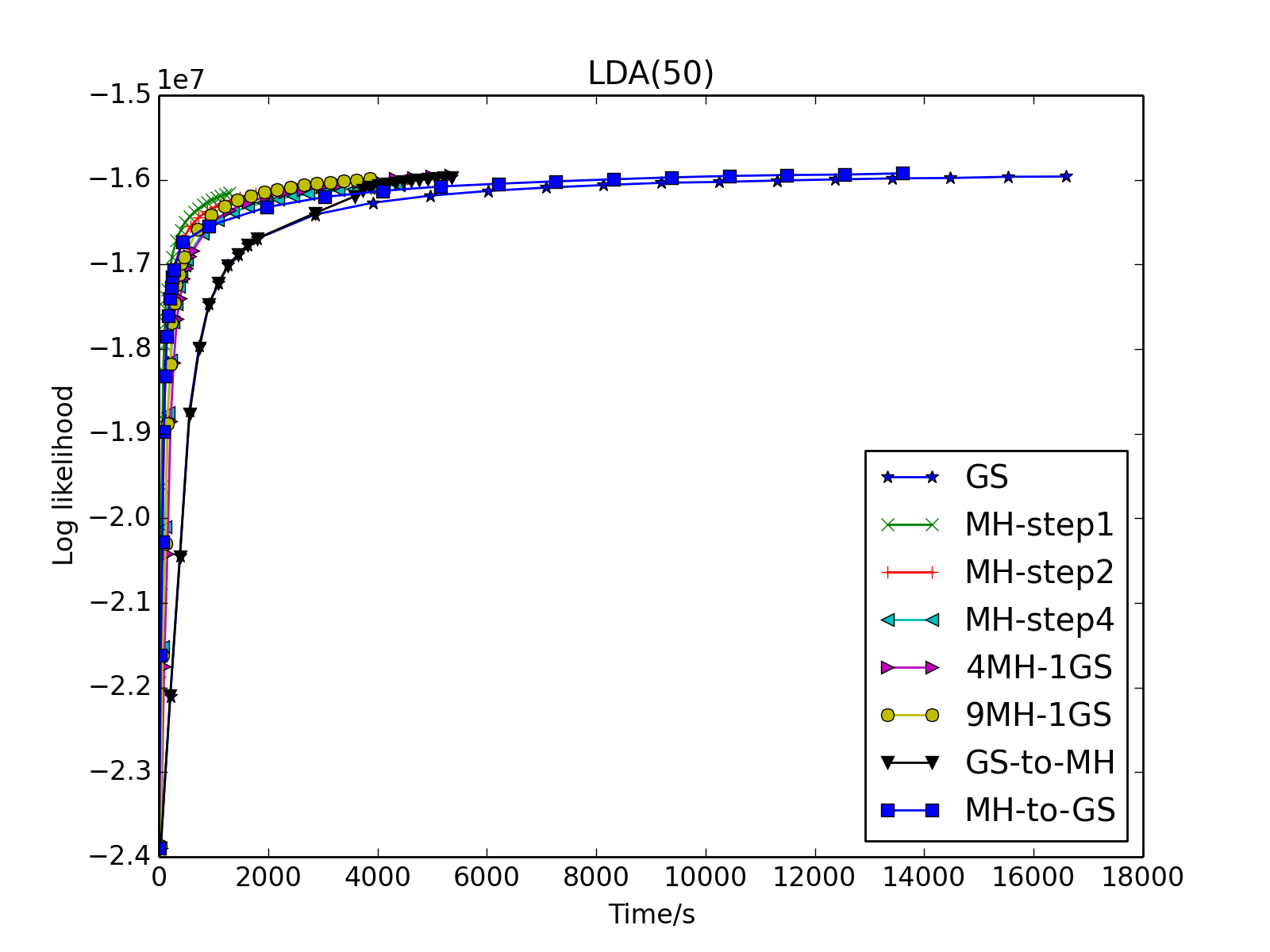} 
}
\subfigure[LDA (K=100)]
{
\includegraphics[width=0.3\textwidth]{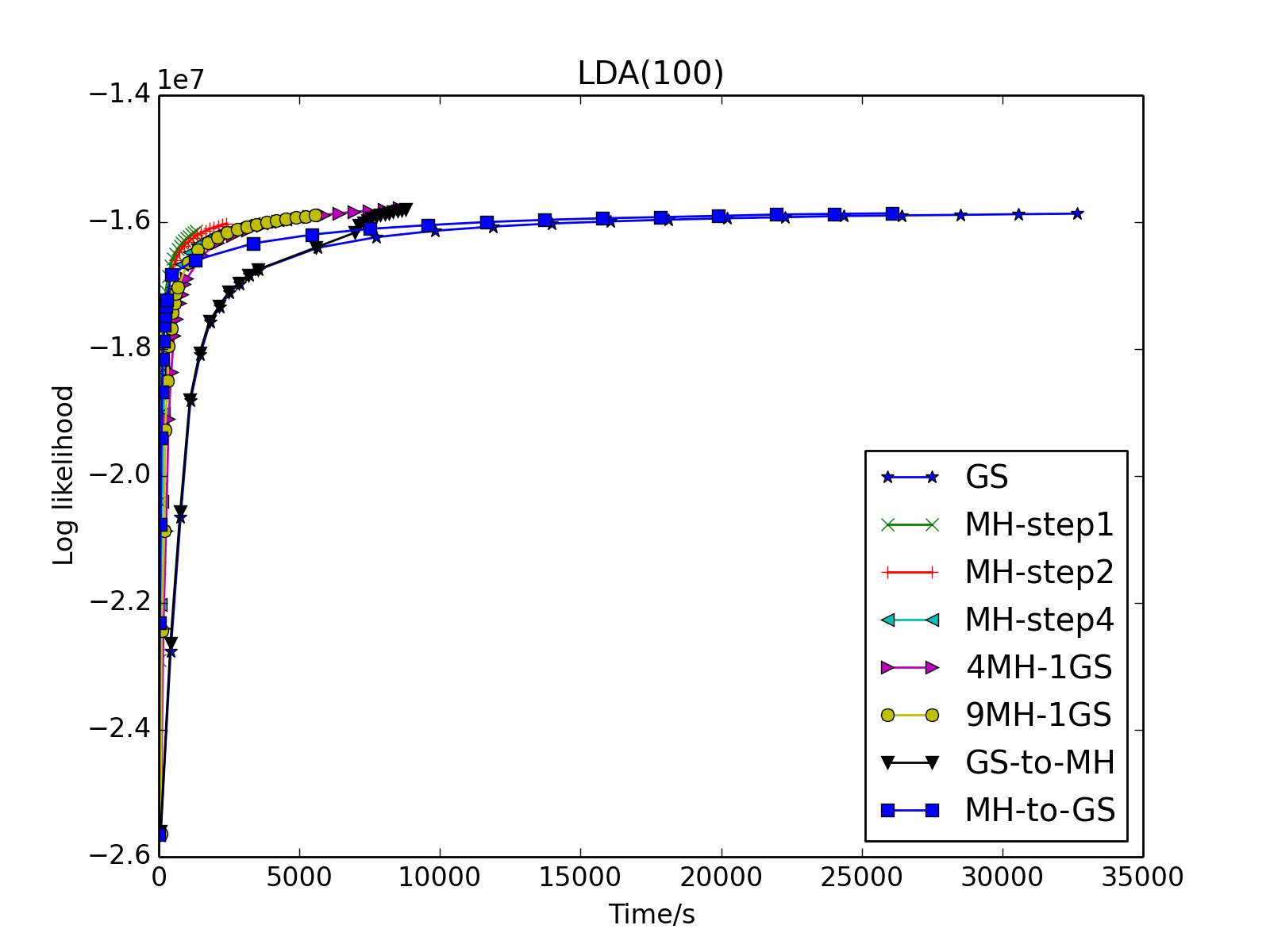} 
}
\quad
\subfigure[Sentence LDA (K=50)]
{
 \includegraphics[width=0.3\textwidth]{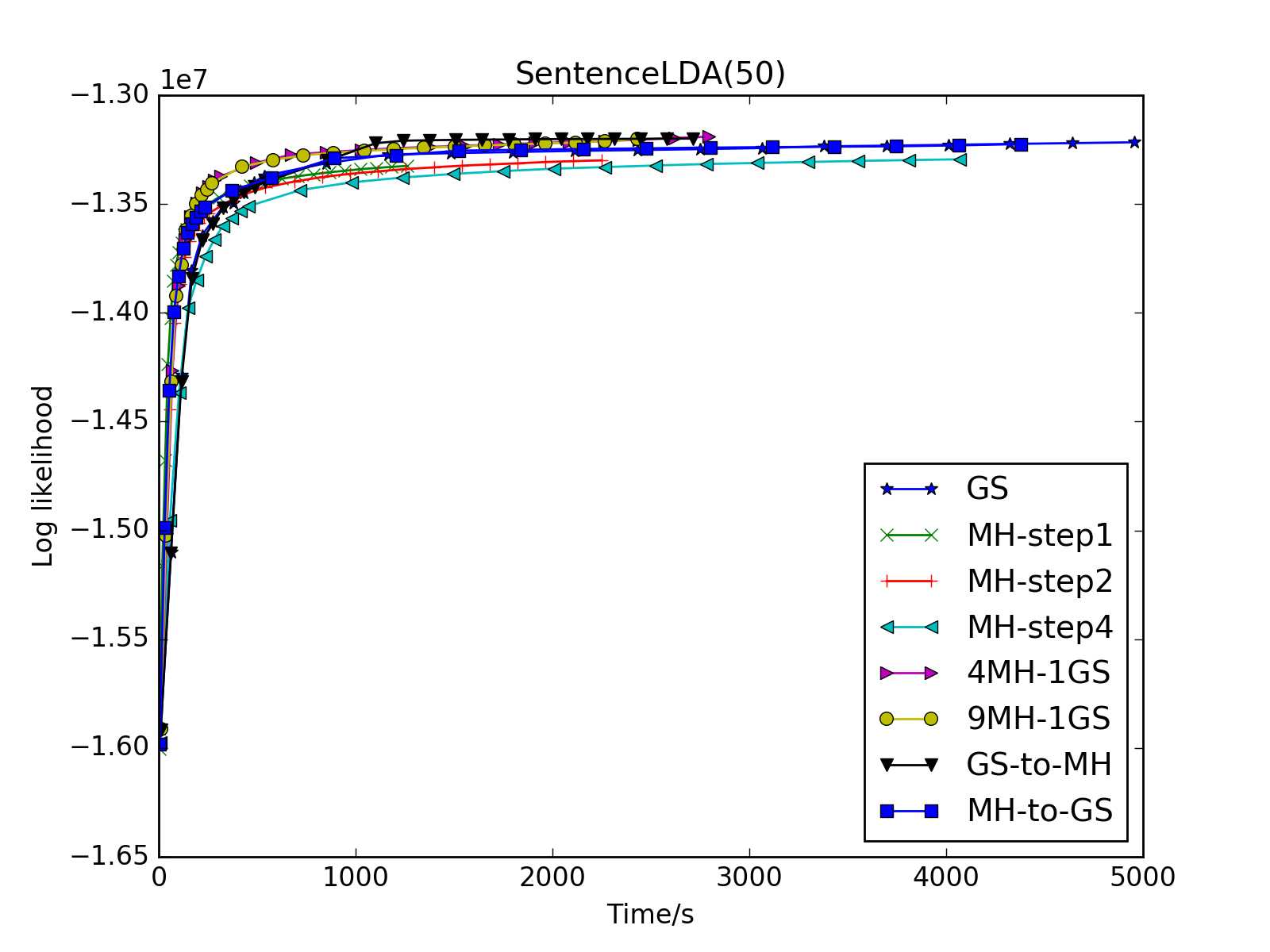} 
}
\subfigure[Sentence LDA (K=100)]
{
\includegraphics[width=0.3\textwidth]{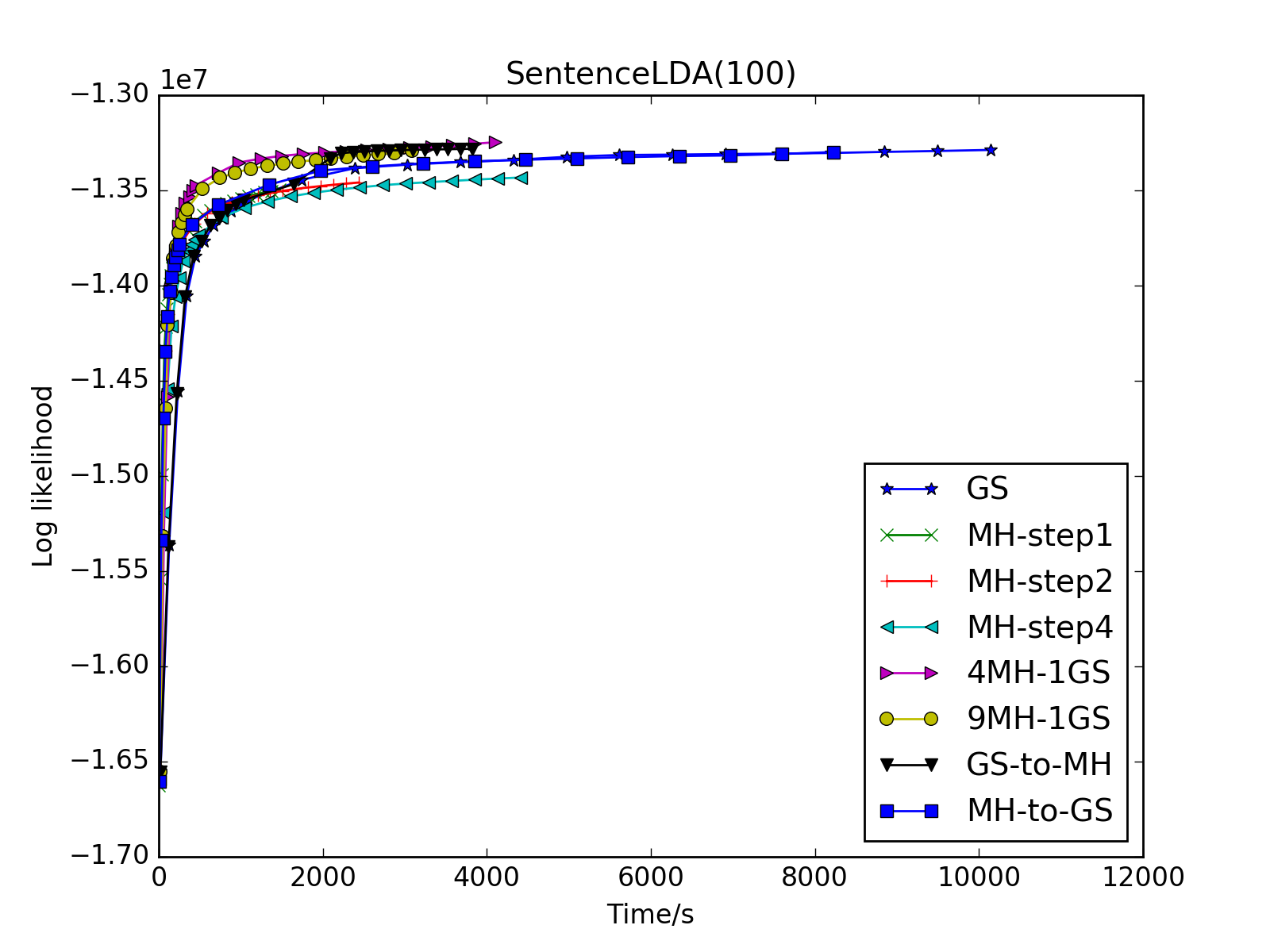} 
}
\quad
\subfigure[TOT (K=50)]
{
 \includegraphics[width=0.3\textwidth]{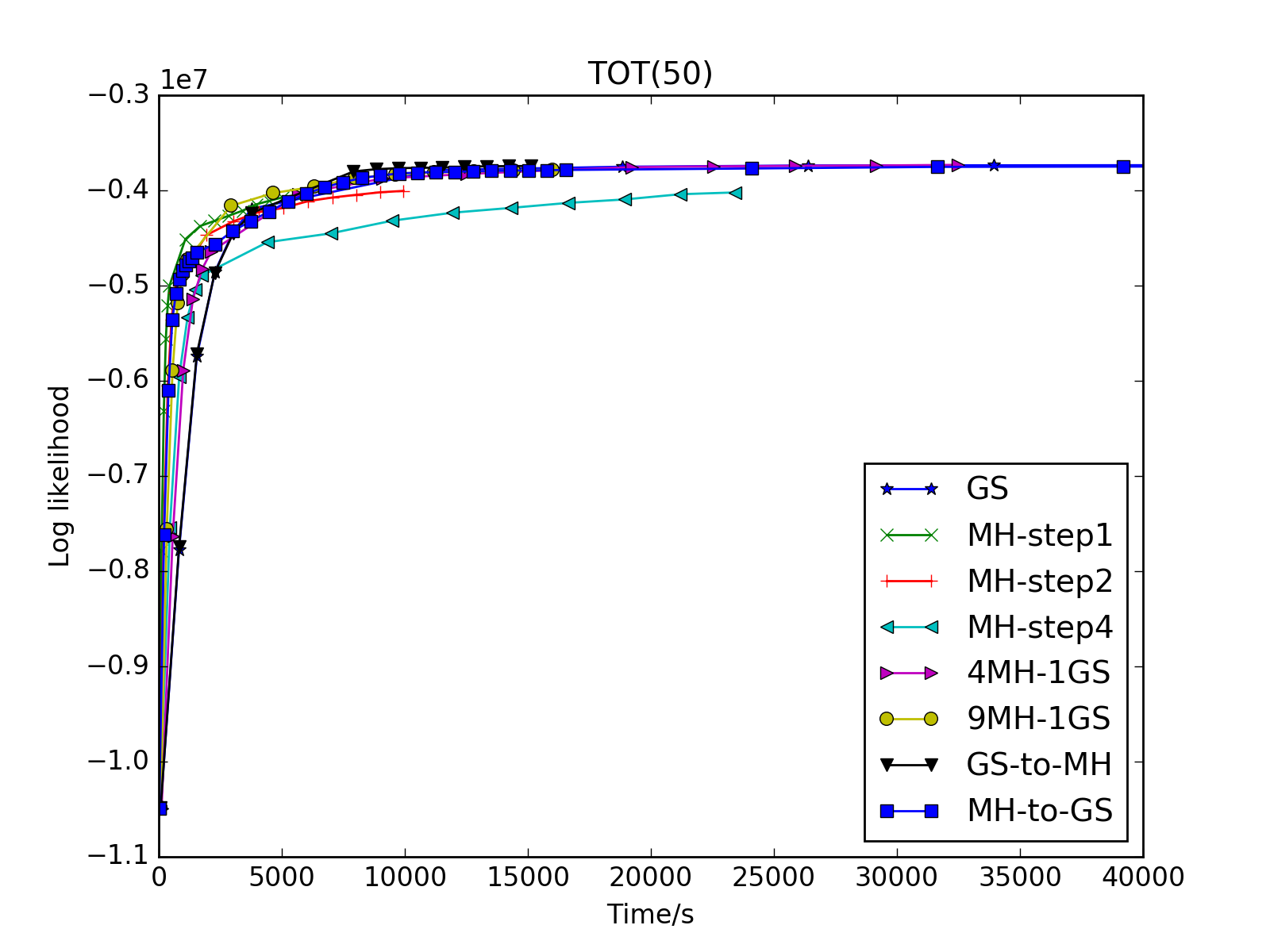} 
}
\subfigure[TOT (K=100)]
{
 \includegraphics[width=0.3\textwidth]{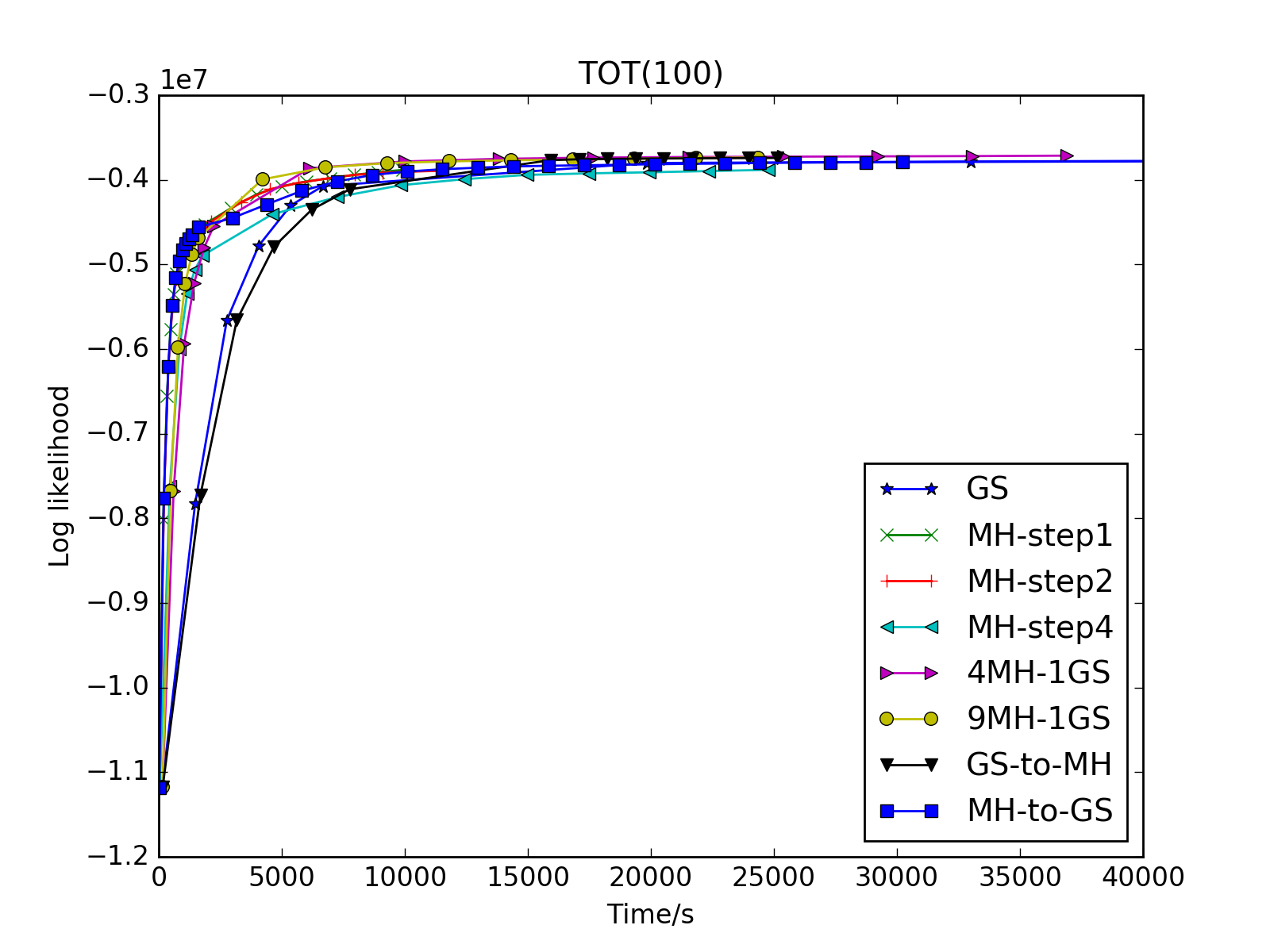} 
}
\caption{Comparison of Sampling Methods (Time) (Best Viewed in Color)}
\label{fig:Comparison of Sampling Methods (Time)}
\end{figure*}

The log likelihood of each sampling methods is plotted against iteration in Figure~\ref{fig:Comparison of Sampling Methods (Iteration)} and against time in Figure~\ref{fig:Comparison of Sampling Methods (Time)}. Therein, MH-step1 is MH with only one MH step. Analogously, MH-step2 and MH-step4 are MH with 2 and 4 MH steps respectively. 4MH-1GS is the sampling method that performs one iteration of GS after every 4 iterations of MH. 9MH-1GS performs one iteration of GS after every 9 iterations of MH. GS-to-MH starts with GS for the first 100 iterations and then switches to MH for the remaining iterations. MH-to-GS starts with MH for the first 100 iterations and then switches to MH for the remaining iterations. Note that space limitation prevents us from presenting more parameter settings for each sampling method. However, the results shown in Figures~\ref{fig:Comparison of Sampling Methods (Iteration)} and \ref{fig:Comparison of Sampling Methods (Time)} are sufficient to showcase our insights obtained from this study.

We first investigate the performance of MH in terms of the number of steps.  From Figure~\ref{fig:Comparison of Sampling Methods (Iteration)}, we observe that larger MH steps usually result in better performance. In most cases, 4-step is better than 2-step and further better than 1-step. The experimental results of all the three topic models verify the above argument. We proceed to compare the performance of different sampling methods.
A possible explanation is that more MH steps help the sampling method to explore more states if the Markov chain has low conductance.

For these three topic models, 4MH-1GS usually achieves the best performance while MH (including all the three MH methods with different steps) usually demonstrates the worst performance. The performance of GS and the other hybrid methods is between 4MH-1GS and MH. 4MH-1GS is better than 9MH-1GS, showing that fairly high frequency of switching MH and GS is effective to improve the model quality.  The hybrid methods have higher chance to prevent the sampling algorithm getting ``stuck'' in a subset of Markov chain states. Hence,  MH is not a good choice if the quality of the resultant model is highly valued.

From Figure~\ref{fig:Comparison of Sampling Methods (Time)}, we observe that MH-step1 achieves a fairly good model within the least time while GS takes the longest time.  The time consumed by hybrid sampling methods is between MH-step1 and GS.  The superiority of MH in efficiency is achieved by reusing the alias tables, which can reduce the amortized time complexity to as low as $O(1)$ per blob. In contrast, the complexity of GS is $O(K)$ per blob, since it needs to calculate the probability for each topic.

Based upon above observations, we obtain the following important insights, which are valid across the three different topic models:
1. GS achieves higher likelihood than MH while MH consumes less time to achieve a fairly good result; 2. Some hybrid sampling methods can achieve even better result than GS while consumes less time than GS. 3. If the quality of the model is the emphasis, hybrid sampling methods like 4MH-1GS should be chosen because it achieves the best model quality with fairly good efficiency. If efficiency is the focus, MH may be chosen since it consumes the least time to generate a reasonably good model.

\subsection{Scalability of Familia}
\label{sec:Scalability of Familia}

\begin{figure*}
\centering
\subfigure[LDA (node=10, K=100)]
{
\includegraphics[width=0.3\textwidth]{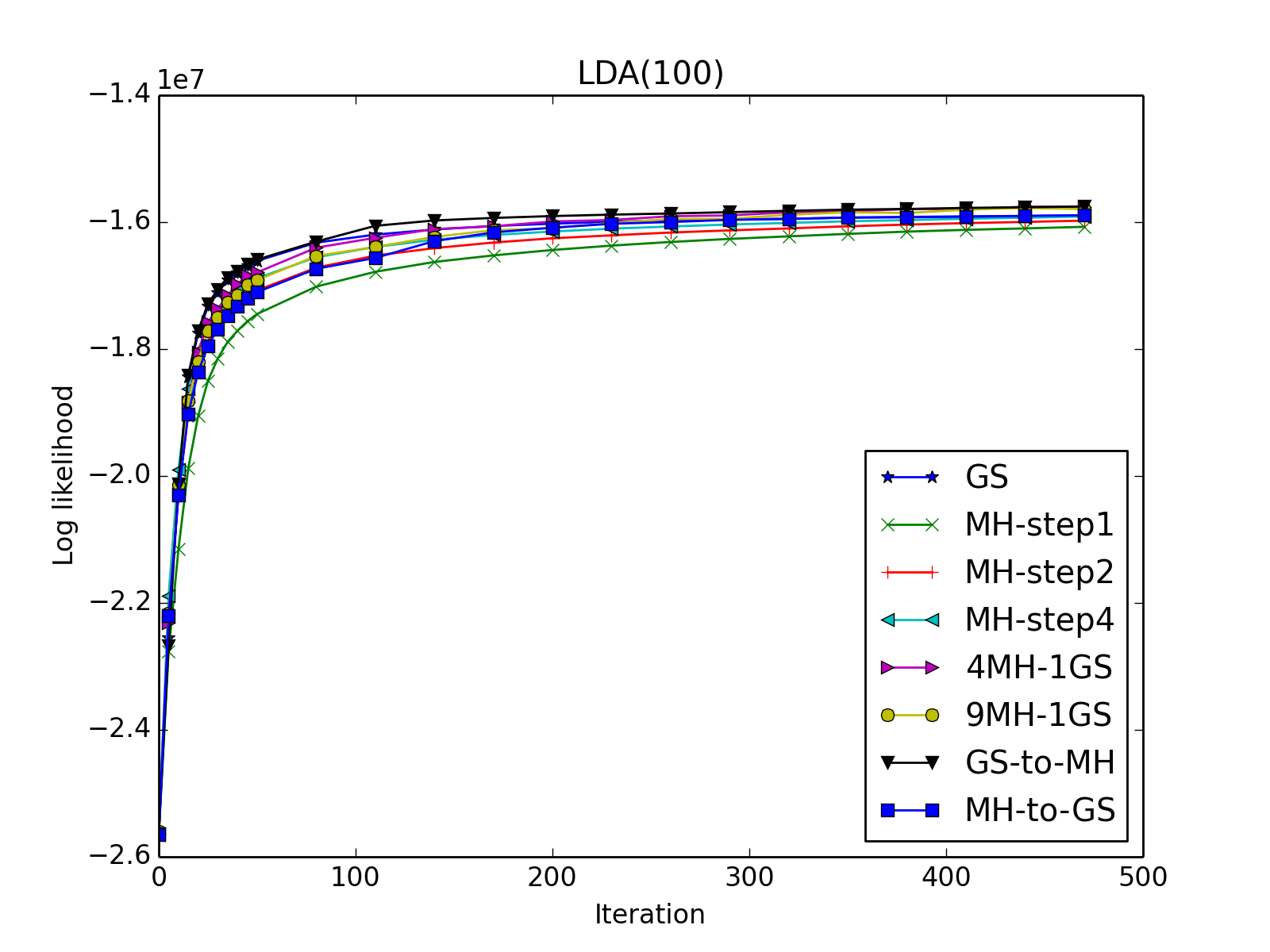}  
}
\subfigure[Sentence LDA (node=10, K=100)]
{
\includegraphics[width=0.3\textwidth]{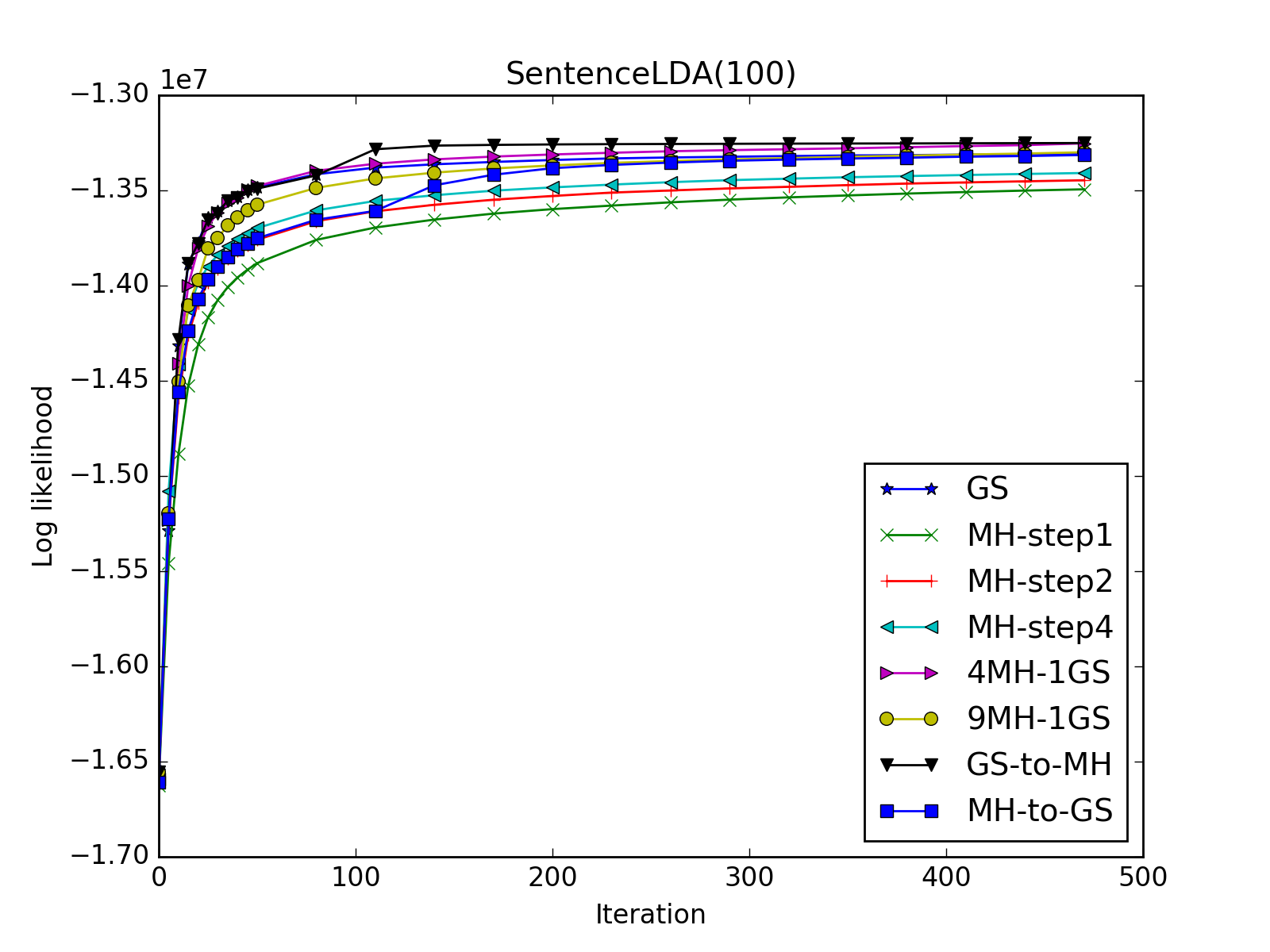} 
}
\subfigure[TOT (node=10, K=100)]
{
 \includegraphics[width=0.3\textwidth]{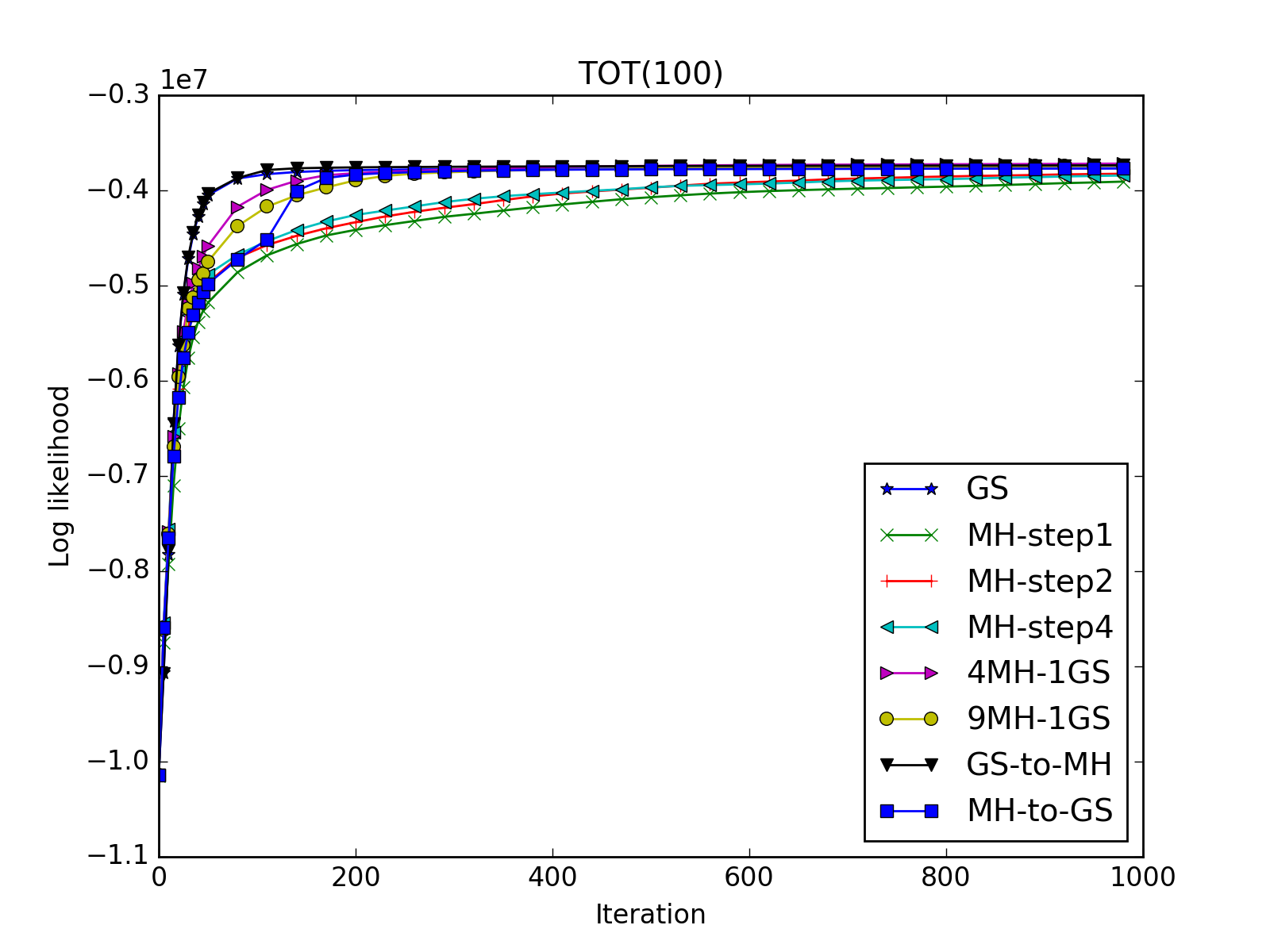} 
}
\caption{Comparison of Sampling Methods (Parallel-Iteration) (Best Viewed in Color)}
\label{fig:Comparison of Sampling Methods (Parallel-Iteration)}
\end{figure*}

\begin{figure*}
\centering
\subfigure[LDA (K=100)]
{
\includegraphics[width=0.3\textwidth]{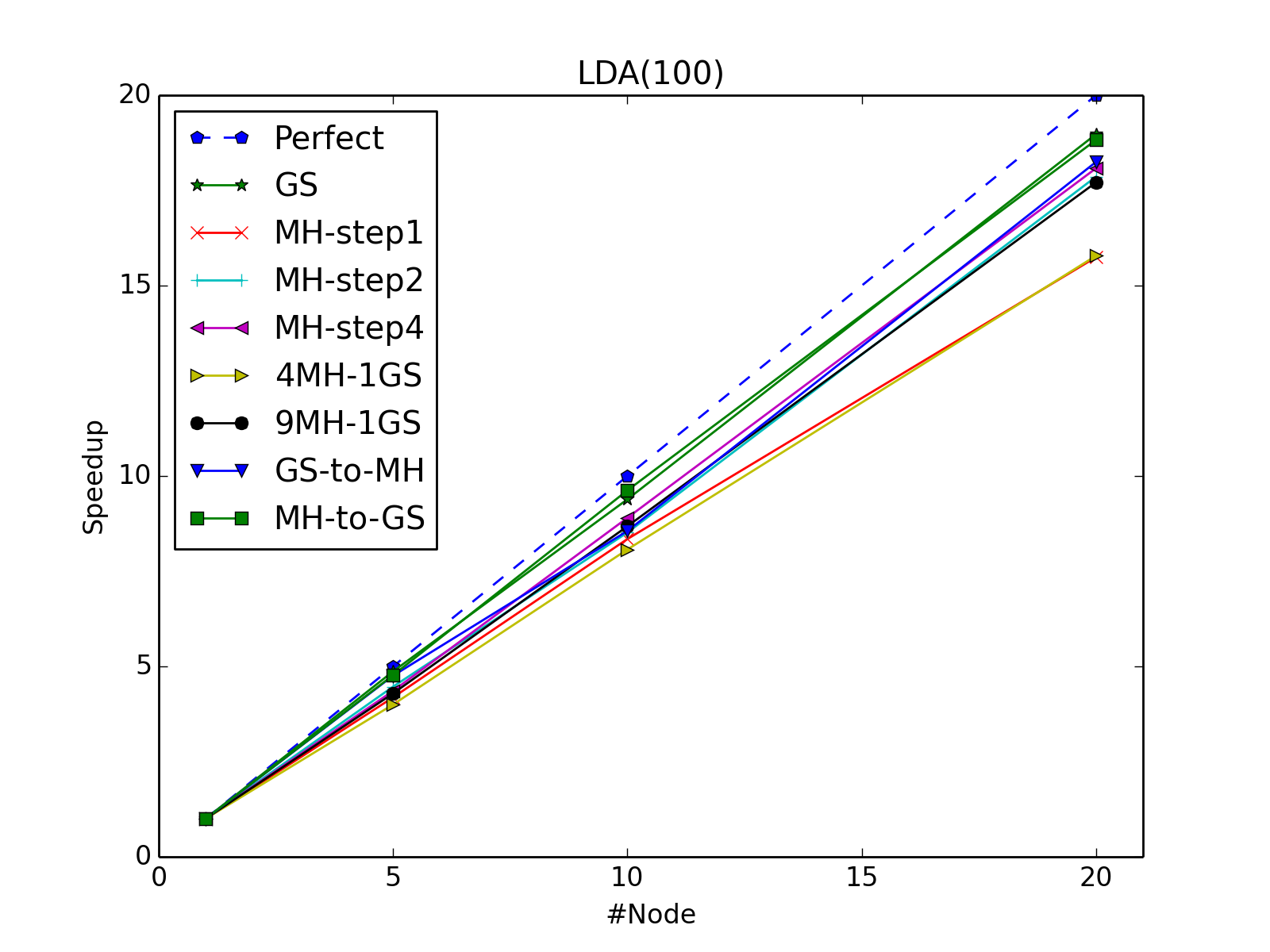}  
}
\subfigure[Sentence LDA (K=100)]
{
\includegraphics[width=0.3\textwidth]{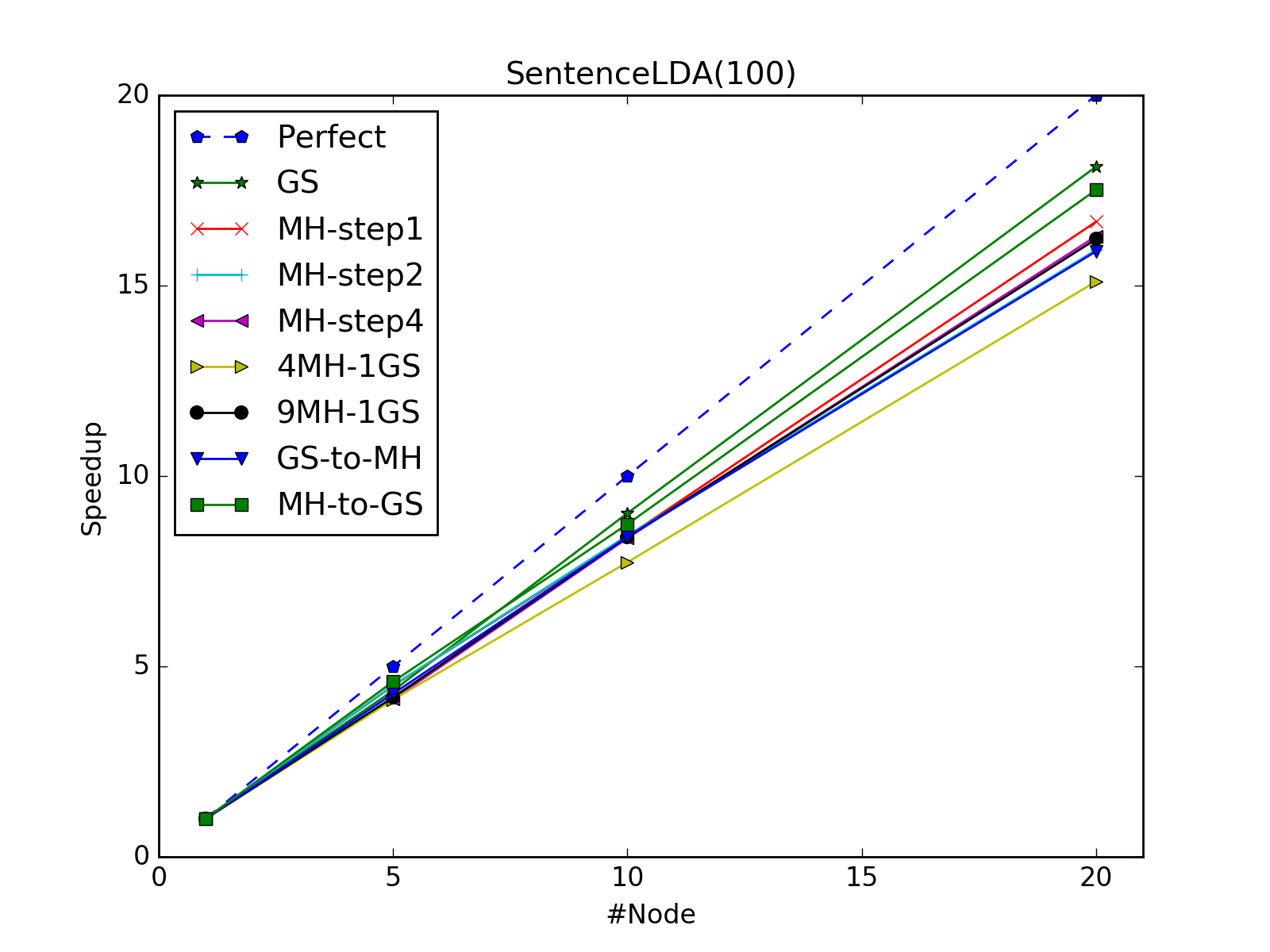} 
}
\subfigure[TOT (K=100)]
{
 \includegraphics[width=0.3\textwidth]{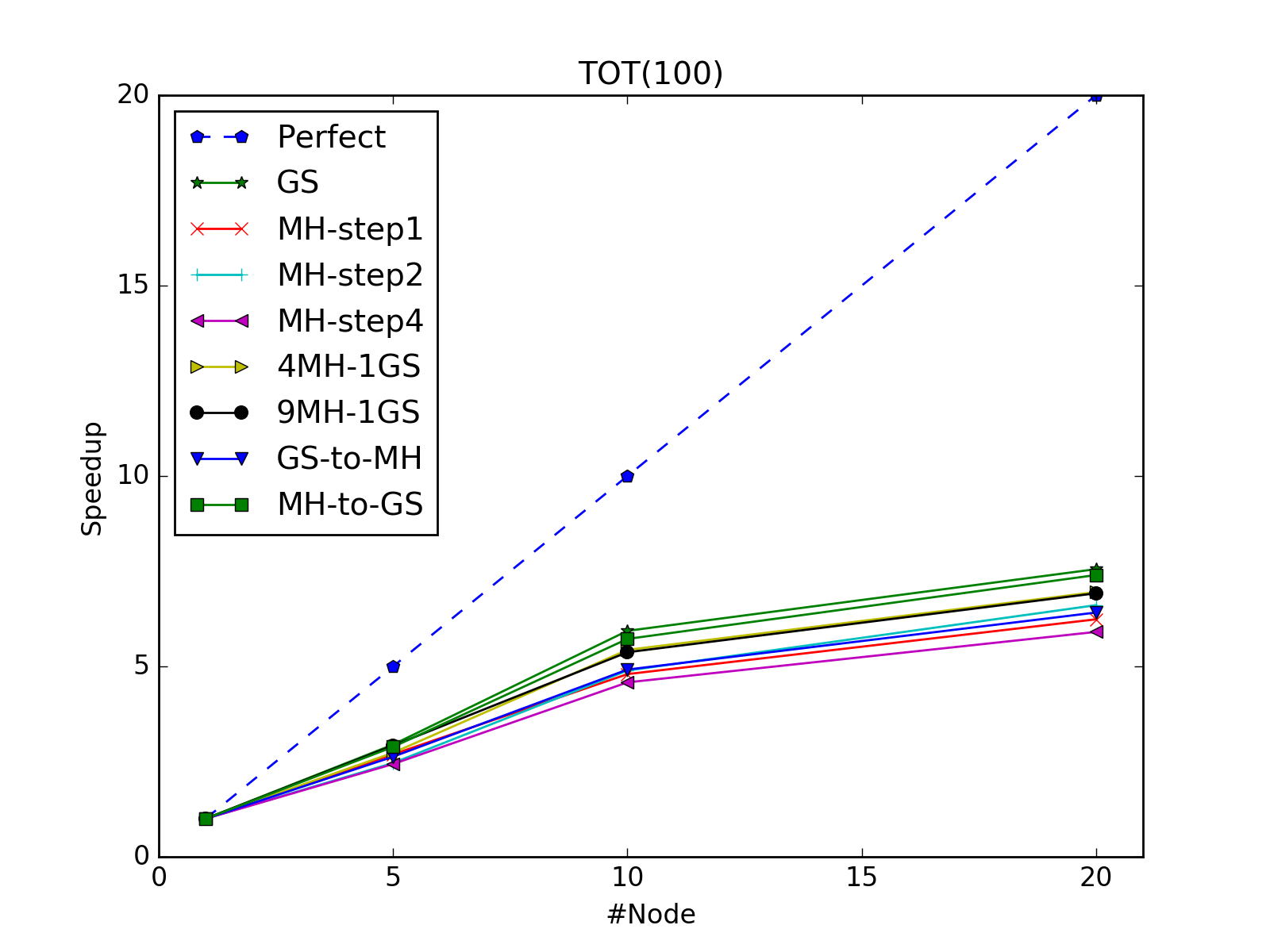} 
}
\caption{Speedup Analysis (Best Viewed in Color)}
\label{fig:Speedup}
\end{figure*}

We proceed to demonstrate the scalability of Familia with 1, 5, 10 and 20 computing nodes. The log likelihood of the three topic models trained on 10 nodes is presented in Figure~\ref{fig:Comparison of Sampling Methods (Parallel-Iteration)}, from which we observe that the results obtained from 10 nodes is aligned with those from a single node, showing that the quality of these three models trained by different sampling methods is not heavily affected by the distributed environment. Although all sampling methods achieve slightly lower log likelihood than those of single node, such degradation is modest in practice. In distributed environment, 4MH-1GS is the method with the best performance and MH methods usually achieve the lowest likelihood. The insights discussed in Section~\ref{sec:Performance of Sampling Methods} still hold for training topic models in distributed environment. Similar phenomenon is observed when the number of nodes is set to 5 or 20 and their results are skipped due to space limitation.

Another important question is how much speedup we obtain when multiple computing nodes are involved. The speedup analysis of the three topic models is presented in Figure~\ref{fig:Speedup}.  High speedup ratio is an indicator of low communication and synchronization cost. With training topic models with PS, low communication cost is primarily achieved by the sparsity of the model under training. The sparser the model is, the less the parameters that each worker needs to pull from servers. When sorted by speedup ratio, the ranking of these sampling methods varies from model to model, showing a specific sampling method has different capability of promoting the sparsity of a topic model. However, for all the three topic models, GS always has the best speedup ratio, indicating that GS is quite effective in promoting the sparsity of the model.  Compared with LDA and Sentence LDA, all sampling methods of TOT have lower speedup ratios due to the synchronization of continuous parameters at each iteration. Hence, topic models without continuous factors can take full advantage of the asynchronous parallelization of PS.  Topic models with continuous factors usually have lower speedup ratio due to the synchronous parallelization caused by continuous parameters.

\section{Industrial Cases and Application}
\label{sec:Industrial Cases}

Currently, Familia has been widely used in both industrial community and academia. In this section, we first propose a guide for users to select appropriate topic models for their tasks and apply them in a proper way using Familia in Subsection~\ref{sec:semantic} and~\ref{sec:matching}. The success of the cases cannot be achieved if LDA is the only topic model in software engineers' arsenal. 
Then a real-life industrial application is presented to further showcase the benefit acquired from the aforementioned paradigms and models using Familia in Subsection~\ref{sec:applications}.

\subsection{Semantic Representation}
\label{sec:semantic}
We first discuss some cases involving semantic representation. The semantic representation derived by topic modeling typically works as features for other machine learning models.

\subsubsection{Document Classification}

\begin{figure*}
\centering
\subfigure[News Topics Distribution as Augmented Features of GBDT]
{
\includegraphics[width=0.3\textwidth]{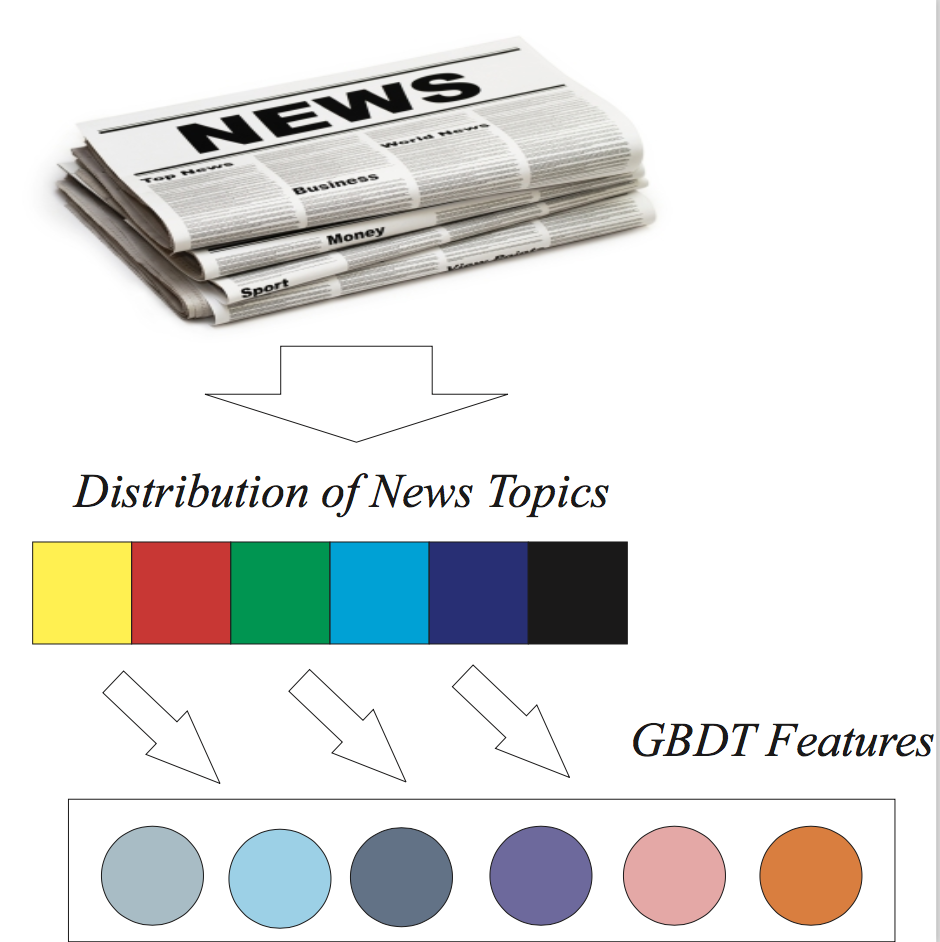}   
\label{pic:news_feature_GDBT}
}
\quad
\subfigure[Experimental Results of News Classification] 
{
\includegraphics[width=0.4\textwidth]{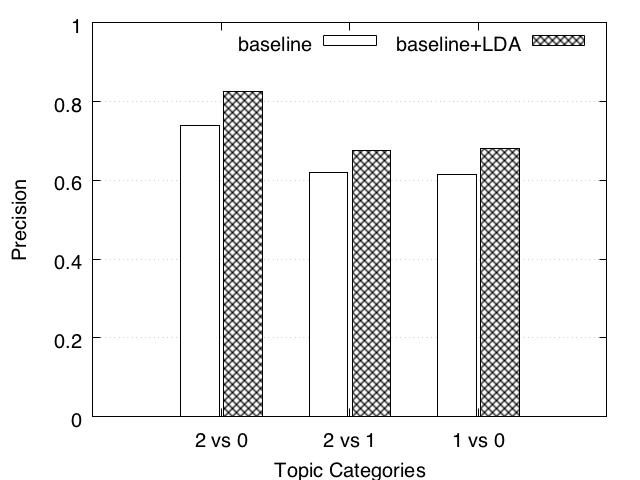}   
\label{pic:news_quality}
}
\caption{Classification of News Articles} 
\label{fig:News Classification}
\end{figure*}

The first case is \textbf{classification of news articles}.
For news feed service, the articles collected from various sources often contain low-quality ones. In order to improve user experience, we need to design a classifier to distinguish the good ones from the bad ones. Conventionally, the classifier is built upon some handcrafted features, which include source sites, text length, the total number of images, etc. We could employ topic model to obtain the topical distribution of each article and augment the handcrafted features with this distribution (shown in Figure \ref{pic:news_feature_GDBT}).  As an experiment, we prepare 7,000 news articles, which are manually labeled into 5 categories, in which 0 stands for those of the lowest quality, and 4 represents for the best. We train Gradient Boosting Decision Tree (GBDT) on 5,000 articles with different features and test the trained classifier on the other 2,000 articles.  Figure~\ref{pic:news_quality} shows the result from the two classifiers using different sets of features: baseline, baseline+LDA. The results of using features of topic model are significantly better, showing that topic model is an effective way for document representation.

\subsubsection{Document Clustering}

\begin{figure}[ht]
\begin{center}
\includegraphics[width=0.7\textwidth]{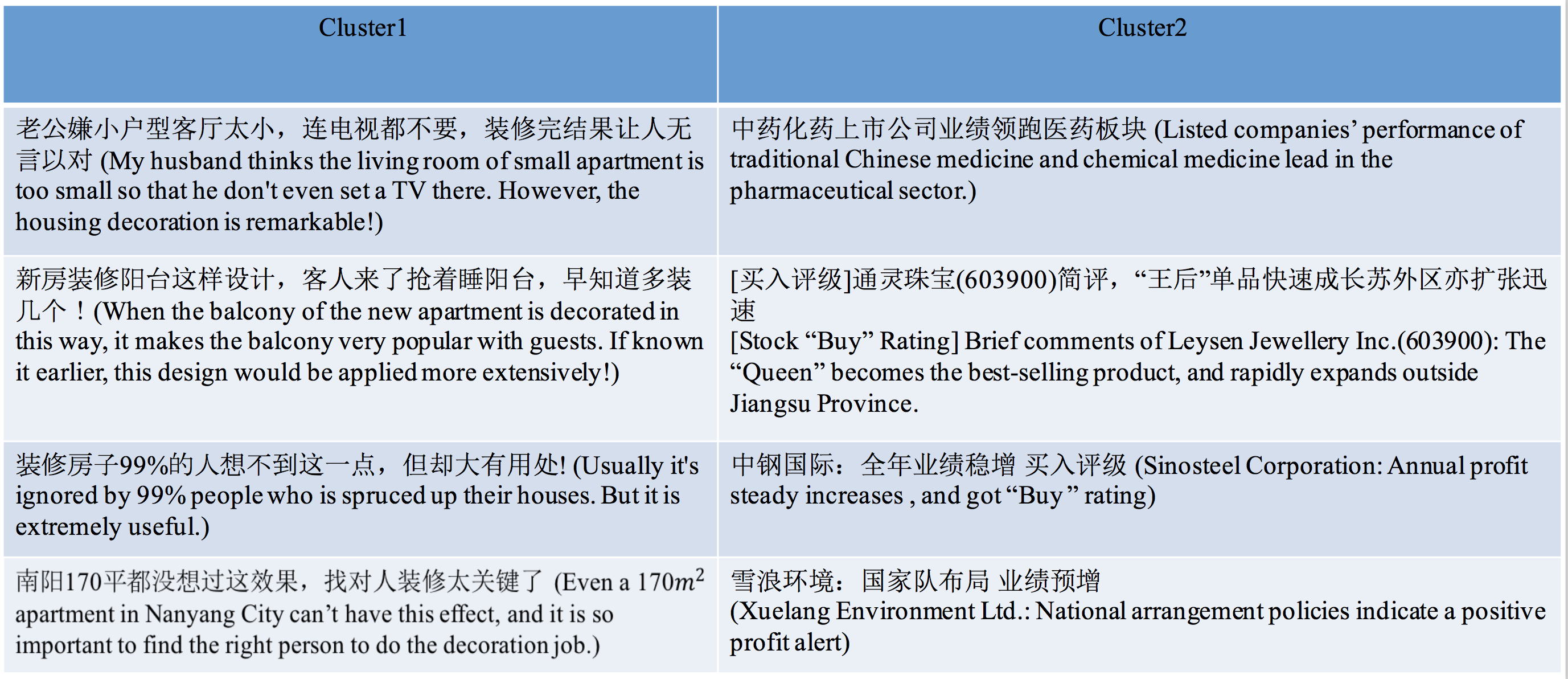}    
\caption{Example of Clustering News Articles}
\label{pic:topic model clustering}
\end{center}
\end{figure}

Straightforwardly, the semantic representation of documents could be utilized for clustering. In the task of \textbf{clustering new articles}, we use LDA to compute the topic distribution of news articles and cluster the articles by K-means. Figure~\ref{pic:topic model clustering} shows two clusters which are obtained by clustering 1000 articles into ten groups. Cluster1 is of articles related to interior design and Cluster2 contains articles about the stock market. The result shows that news articles can be semantically clustered based on their topic distributions. 

\subsubsection{Dimensionality Reduction in News Quality Evaluation}

\begin{figure}[htb]
  \begin{center}
    \includegraphics[width=0.5\textwidth]{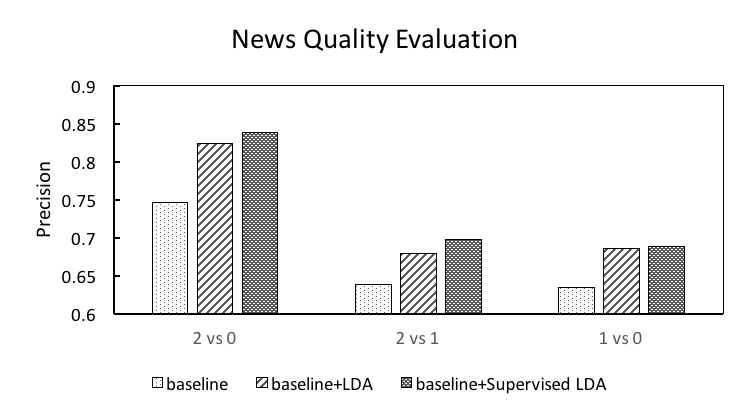}  
    \caption{News Quality Evaluation}
    \label{pic:news quality}
  \end{center}
\end{figure}

Quality evaluation is critical for news recommendation. We now show how topic distributions of each news articles are utilized for news quality evaluation. The topic distribution of each news article is added as extra features for Gradient Boosting Decision Tree \cite{friedman2002stochastic}. We utilize one-day news articles collected by a commercial spider for training and 2000 news articles for testing. Each article is labeled by human experts with a quality indicator ranging from 0 to 2, where 0 indicates a poor quality while 2 shows that the article is of high quality.  We compare different feature settings (i.e., baseline, baseline+LDA and baseline+Supervised LDA) in terms of the precision of separating news from different quality category. The baseline is composed of statistical features such as article length, image amount, entity amount, etc. Two insights are obtained from the experimental result shown in Figure~\ref{pic:news quality}: (1) The lower-dimensional topic representation is a good feature for article quality evaluation and it is effective in boosting the performance; (2) Supervised LDA outperforms LDA since it incorporates quality signals in its generative process and the resultant topic space is more effective for this particular task.

\subsection{Semantic Matching}
\label{sec:matching}
Another paradigm is semantic matching, which can be further categorized as  short-short text matching, short-long text matching and long-long text matching. 

\subsubsection{Short-Short Text Matching}

The need for short-short text matching is common in web search, where we need to compute the semantic similarity between queries and web page titles. Due to the difficulty of topic modeling on short text, embedding-based models such as Word2Vec and Topical Word Embeddings (TWE) are much more common for this task. Assume we want to compute the semantic similarity between a query $q=$``recommend good movies'' and a web page title $t=$``2016 good movies in China'', we first convert the queries into their embeddings (i.e., $\vec{q}$ and $\vec{t}$) and then compute the semantic similarity between these embeddings with the metric of cosine similarity.

\begin{equation}
\label{eq:cos_sim}
\begin{aligned}
CosineSimilarity(\vec{q}, \vec{t}) = \frac{\vec{q} \cdot \vec{t}}{|\vec{q} | |\vec{t}|}
\end{aligned}
\end{equation}

\noindent There are more sophisticated short-short text matching mechanisms in literature, interested readers may refer to deep neural network based models such as Deep Structured Semantic Model (DSSM) \cite{huang2013learning} and Convolutional Latent Semantic Model (CLSM) \cite{shen2014latent}.

\subsubsection{Short-Long Text Matching}

In many online applications, we need to compute the semantic similarity between query and document. Since query is typically short and document content is much longer, short-long text matching is needed in this scenario.  Due to the difficulty of topic inference on short text, we compute the probability of the short text generated from the topic distribution of the long text as follows:

\begin{equation}
\label{eq:cos-sim}
\begin{aligned}
Similarity(q,c)=\prod_{w\in q}\sum_{k}P(w|z_k)P(z_k|c),
\end{aligned}
\end{equation}
where $q$ stands for query, $c$ for document content, $w$ for words in query and $z_k$ for topics.

\begin{figure}
\centering
\subfigure[Baseline: An ad about Love Story Micro-movie]
{
\includegraphics[width=0.48\textwidth]{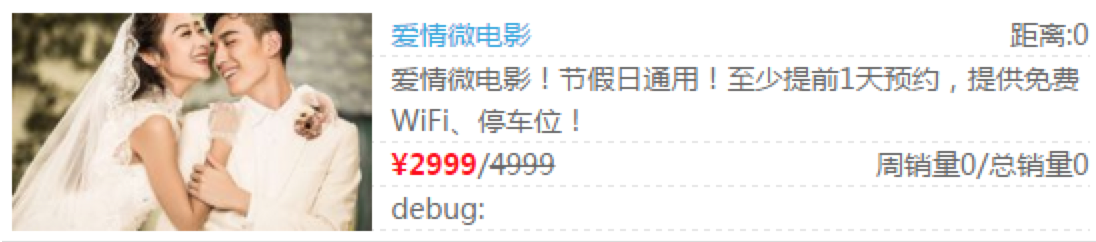}  
}
\subfigure[Result with SentenceLDA Feature: An ad about Aijia Wedding Photography]
{
\includegraphics[width=0.48\textwidth]{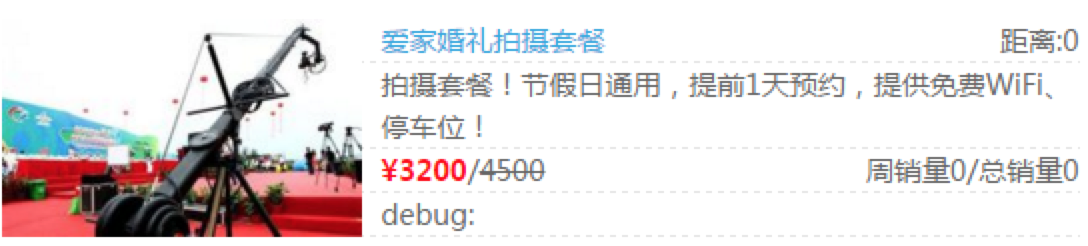}  
}
\caption{Semantic Matching of Query-Ad}
\label{fig:user-webpage}
\end{figure}

We first discuss the task of \textbf{online advertising}, in which we need to compute the semantic similarity between query and ad pages. We treat each textual field on ad page as a sentence and apply SentenceLDA for this task. After obtaining the topic distribution of each ad page,  we apply Eq.(\ref{eq:cos-sim}) to compute the semantic similarity between query and the ad page. Such similarity can be utilized as a feature in downstream ranking models. For a query ``recording of wedding ceremony'', we compare its ranking results from two strategies in Figure \ref{fig:user-webpage}. We can see that the result with SentenceLDA feature is better at satisfying the underlying need of the query.

\begin{figure}[h]
\begin{center}
\includegraphics[width=0.4\textwidth]{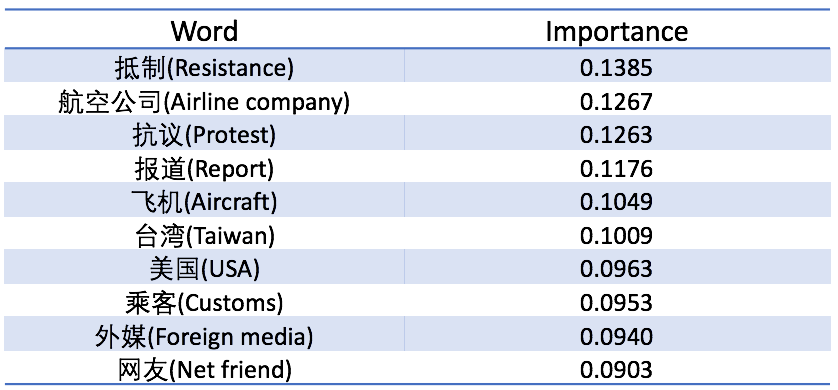} 
\caption{Keyword Extraction based on TWE}
\label{pic:keyword}
\end{center}
\end{figure}

An extreme case of short-long text matching is the task of \textbf{keyword extraction from document}.  We extract a set of keywords from documents as concise and explicit representation of the document. The conventional way of extracting keywords from texts relies upon the TF and IDF information. If we want to introduce the semantic importance, we can use Eq.(\ref{eq:TWE keyword}) to compute the similarity of a word and the document as follows:

\begin{equation}
\label{eq:TWE keyword}
\begin{aligned}
Similarity(w, c) = \sum_{k=1}^K{cos(\vec {v_w}, \vec {z_k})P(z_k|c)},
\end{aligned}
\end{equation}
where $c$ stands for document content, $w$ for each word, $\vec{v_w}$ for word embedding for word $w$ and $\vec{z_k}$ for vector representation of topic $z$. 
We use Eq.(\ref{eq:TWE keyword}) to compute the similarity between each word and the whole article. Top-10 keywords (with stop words eliminated) extracted by TWE are shown in Figure \ref{pic:keyword}, and we can see that the keywords from TWE preserve the important information in the news.

\subsubsection{Long-Long Text Matching}

We can evaluate the semantic similarity between long texts by the distance of their topical distributions. Such semantic similarity can be further utilized as a feature in various machine learning models.  The distance metrics of gauging two topical distributions include Hellinger Distance (HD) and Jensen-Shannon Divergence (JSD). Hellinger Distance is formally defined as follows:

\begin{equation}
\small
HD(P,Q) = \frac{1}{\sqrt{2}}\sqrt{\sum_{i=1}^{K}(\sqrt{p_{i}}-\sqrt{q_{i}})^{2}},
\end{equation}
where $p_{i}$ and $q_{i}$ are the $i$-th element of the corresponding distributions. The definition of Jensen-Shannon Divergence (JSD) is as follows:
\begin{equation}
\begin{aligned}
\small
JSD(P||Q) = \frac 1 2 (KLD(P||M) + KLD(Q||M))
\end{aligned}
\end{equation}
\begin{equation}
\small
\begin{aligned}
M = \frac 1 2 (P + Q)
\end{aligned}
\end{equation}
\begin{equation}
\small
\begin{aligned}
KLD(P||M) = \sum_{i=1}^K p_i \ln \frac {p_i} {m_i}
\end{aligned}
\end{equation}
where $KLD$ stands for Kullback-Leibler Divergence. 

\subsection{Application}
\label{sec:applications}

We proceed to discuss the task of \textbf{personalized fiction recommendation}. Matrix factorization is a common approach for industrial recommendation systems. SVDFeature \cite{chen2012svdfeature} is a framework designed to efficiently solve the feature-based matrix factorization. SVDFeature is quite flexible and is able to accommodates global features, user features and item features. SVDFearure can be mathematically described as follows:

\begin{equation}
\small
\begin{aligned}
\label{eq:svdfeature}
y = \mu + (\sum_{j}b_{j}^{(g)}\gamma_{j} + \sum_{j}b_{j}^{(u)}\alpha_{j} + \sum_{j}b_{j}^{(i)}\beta_{j}) + (\sum_{j}p_{j}\alpha_{j})^{T}(\sum_{j}q_{j}\beta_{j}),
\end{aligned}
\end{equation}
where $y$ is target, $\mu$ is a constant indicating the global mean value of target, $\alpha$ represents user feature, $\beta$ represents item feature, $\gamma$ represents global feature, $b^{(g)}$ is weight of global feature, $b^{(u)}$ is weight of user feature, $b^{(i)}$ is weight of item feature, $p$ and $q$ are model parameters.

\begin{figure}[h]
\begin{center}
\includegraphics[width=0.6\textwidth]{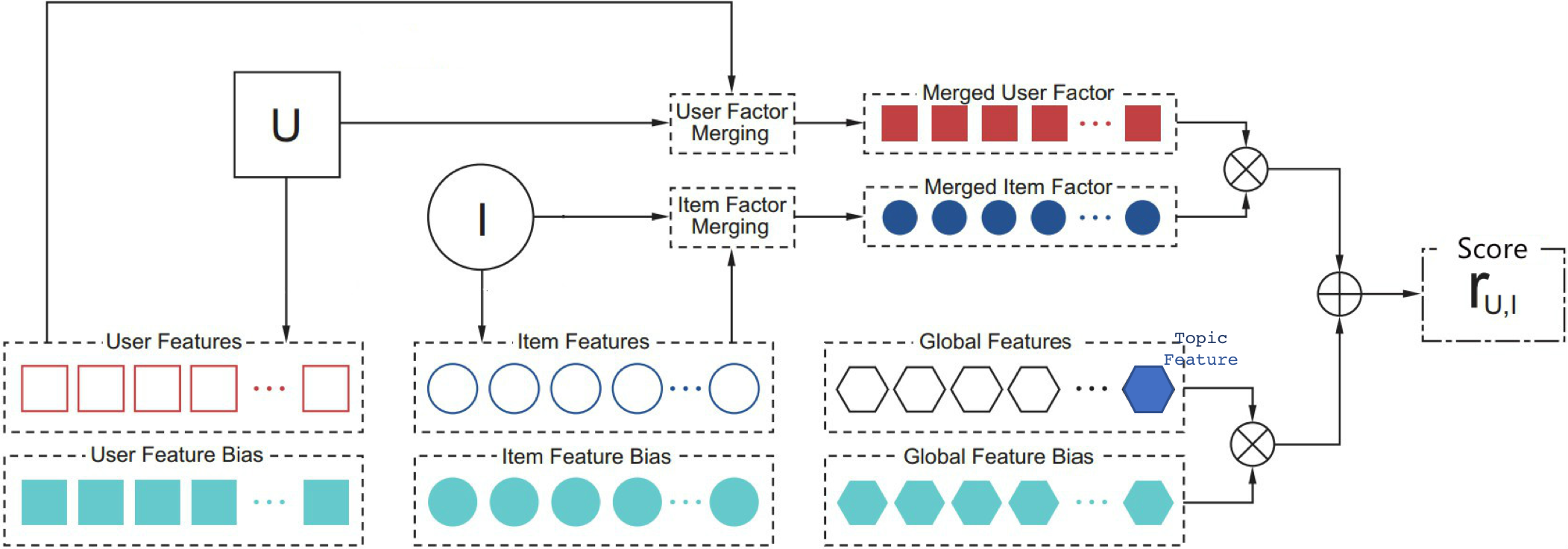} 
\caption{SVDFeature with Topic Feature}
\label{pic:feature-based}
\end{center}
\end{figure}

\begin{figure}[htp]
  \centering
  \includegraphics[width=0.5\textwidth]{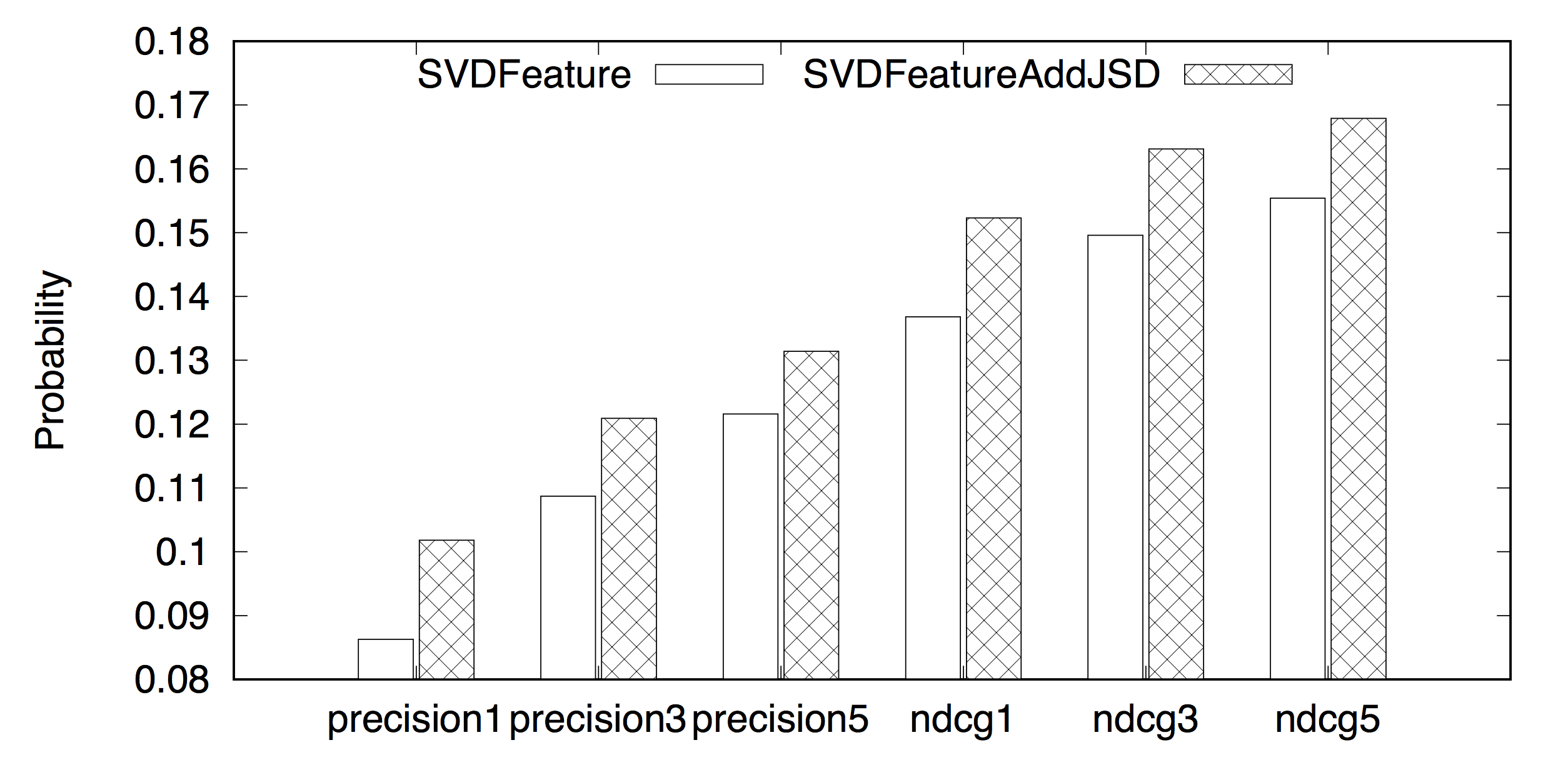} 
  \caption{Fiction Recommendation Performance}
  \label{pic:Fiction Recommendation Performance}
\end{figure}

In the scenario of personalized fiction recommendation, each user has some historically downloaded fictions. By conduct topic modeling on these fictions, we can obtain the user's topic representation, which works as a user profile of reading interests. By computing the JSD between the topic distribution of each fiction and the user profile, we can quantify the probability that user is interested in this fiction. We augment the aforementioned SVDFeature framework with the JSD value as a global feature (Figure \ref{pic:feature-based}). From a comparative study shown in Figure~\ref{pic:Fiction Recommendation Performance}, we can see that adding JSD is effective to improve the performance of SVDFeature. SVDFeature with JSD constantly outperform its original counterpart in terms of both Precision and NDCG.

\section{Conclusion}
\label{sec:Conclusion}

In this paper, we propose a configurable topic modeling framework named Familia for industrial text engineering. The framework provides novel functionalities such as topic model customization, automatic parameter inference and post-modeling utilities.  Based on the hybrid sampling mechanism of Familia, we further provide practical suggestions of choosing proper sampling methods for different topic models. Equipped with Familia, software engineers can easily test different assumptions of the latent structure of their data without tediously deriving mathematical equations and implementing sampling algorithms from scratch. We wish that Familia would help the technique of topic modeling to be utilized in more proper and convenient manner in industrial scenarios.

\bibliographystyle{ACM-Reference-Format-Journals}
\bibliography{familia}

\end{document}